%% file: acl_latex.tex
\documentclass[11pt]{article}

\usepackage[preprint]{acl}

\usepackage{times}
\usepackage{latexsym}

\usepackage[T1]{fontenc}

\usepackage[utf8]{inputenc}

\usepackage{microtype}

\usepackage{inconsolata}
\usepackage{listings}

\usepackage{graphicx}

\usepackage[most]{tcolorbox}

\newtcblisting[auto counter]{promptbox}[2][]{%
    enhanced,
    listing only,
    listing engine=listings,
    width=\linewidth,
    halign=left,
    colback=white,
    colframe=black!45,
    colbacktitle=black!10,
    coltitle=black,
    fonttitle=\bfseries\small,
    title={Prompt~\thetcbcounter: #2},
    boxrule=0.5pt,
    arc=1mm,
    left=7pt,
    right=7pt,
    top=6pt,
    bottom=6pt,
    listing options={
        basicstyle=\ttfamily\small,
        columns=fullflexible,
        keepspaces=true,
        breaklines=true,
        breakindent=0pt,
        showstringspaces=false,
        xleftmargin=0pt,
        aboveskip=0pt,
        belowskip=0pt,
    },
    #1
}

\usepackage{xcolor}

\newtcblisting[auto counter]{codebox}[2][]{%
    enhanced,
    listing only,
    listing engine=listings,
    width=\linewidth,
    colback=white,
    colframe=black!40,
    colbacktitle=black!10,
    coltitle=black,
    fonttitle=\bfseries\small,
    title={#2},
    boxrule=0.5pt,
    arc=1mm,
    left=7pt,
    right=7pt,
    top=6pt,
    bottom=6pt,
    listing options={
        language=Python,
        basicstyle=\ttfamily\footnotesize,
        keywordstyle=\color{blue!65!black}\bfseries,
        commentstyle=\color{gray!75!black}\itshape,
        stringstyle=\color{green!45!black},
        emph={MemoClass,Basic_Recorder,Dict,List,Union},
        emphstyle=\color{teal!75!black},
        morekeywords={async,await},
        columns=fullflexible,
        keepspaces=true,
        breaklines=true,
        breakindent=0pt,
        showstringspaces=false,
        aboveskip=0pt,
        belowskip=0pt,
    },
    #1
}

\title{Discovering Effective Agent Memory Designs through Program Evolution}

\author{
  Xinheng Ding\textsuperscript{1}, Seyyed Saeid Cheshmi\textsuperscript{1}, Ziyang Zhang\textsuperscript{1}, Tairan Chen\textsuperscript{2} \\
  \textbf{Wenya Xie\textsuperscript{1}, Yu Hou\textsuperscript{1}, Rui Zhang\textsuperscript{1}, Zirui Liu\textsuperscript{1}} \\
  \textsuperscript{1}University of Minnesota \\
  \textsuperscript{2}Stevens Institute of Technology
}

\newcommand{\sysname}[1][]{\textnormal{\fontfamily{cmtt}\selectfont MAPE#1}}   %
\newcommand{\mono}[1]{\textnormal{\fontfamily{cmtt}\selectfont #1}}   %
\usepackage{pifont}
\newcommand{\circlednum}[1]{\ding{\numexpr181+#1\relax}}

\usepackage{xcolor}
\usepackage{tabularx}
\usepackage{placeins}
\usepackage{amsmath}
\usepackage{booktabs}
\usepackage{colortbl}
\usepackage{multirow}
\usepackage{array}
\usepackage{fontawesome5}
\usepackage{tikz}
\usetikzlibrary{positioning, arrows.meta, fit, calc, backgrounds}

\begin{document}
\maketitle
\begin{abstract}
\input{sections/abstract}
\end{abstract}

\input{sections/introduction}

\input{sections/method}

\input{sections/tab_main}

\input{sections/experimental_setup}

\section{Results and Analysis}
\label{sec:results}
\input{sections/fig_memorph_a}

\subsection{Performance Evaluation}
\label{ssec:overall_results}

On the LoCoMo evaluation set with GPT-4.1 mini, \sysname[-A\textsubscript{LCM}] and \sysname[-P\textsubscript{LCM}] achieve overall accuracies of 87.94\% and 85.18\%, respectively, as shown in Table~\ref{tab:locomo}. \sysname[-A\textsubscript{LCM}] outperforms all baselines across every question category, while \sysname[-P\textsubscript{LCM}] retains competitive accuracy at approximately 28\% of its counterpart's lifecycle serving cost. On LongMemEval with Qwen3-30B-A3B-Instruct-2507, \sysname[-A\textsubscript{LME}] and \sysname[-P\textsubscript{LME}] achieve 77.00\% and 72.25\% accuracy, respectively, both exceeding RAG, the strongest baseline at 66.00\%. \sysname[-A\textsubscript{LME}] achieves an absolute accuracy gain of 11.00\% over RAG, as shown in Table~\ref{tab:longmemeval}.

Replacing the serving backbone of the LoCoMo-evolved programs, including the QA model, with Qwen3-30B-A3B-Instruct-2507 preserves strong performance without further evolution. Under this backbone, \sysname[-P\textsubscript{LCM}] achieves 81.33\% accuracy, an absolute gain of 12.01\% over SimpleMem at 35\% of its serving cost. The ranking between the two evolved programs reverses, with \sysname[-A\textsubscript{LCM}] achieving 78.58\%. This reversal is driven by temporal questions: \sysname[-A\textsubscript{LCM}] drops from 79.84\% to 43.80\%, whereas \sysname[-P\textsubscript{LCM}] retains 70.93\%. \sysname[-A\textsubscript{LCM}] leaves the final date-resolution step to the QA model, while \sysname[-P\textsubscript{LCM}] explicitly converts relative time expressions into absolute dates before answering, which may help preserve temporal accuracy across backbones. A concrete example is illustrated in Figure~\ref{fig:temporal_example}.

We further evaluate the selected programs on the other benchmark without additional evolution, using the fixed input adapters described in Appendix~\ref{app:transfer}. As shown in Table~\ref{tab:longmemeval_transfer}, \sysname[-A\textsubscript{LCM}] and \sysname[-P\textsubscript{LCM}] achieve 65.50\% and 64.25\% accuracy on LongMemEval, respectively. Both transferred programs outperform the five evaluated memory systems, while RAG achieves 66.00\% accuracy. In the other direction, \sysname[-A\textsubscript{LME}] and \sysname[-P\textsubscript{LME}] achieve 75.90\% and 74.67\% accuracy on LoCoMo, respectively, exceeding all baselines in Table~\ref{tab:locomo_transfer}. These results support the transferability of the discovered memory designs beyond their search benchmarks, suggesting that their gains are not solely attributable to benchmark-specific adaptation.

\subsection{Comparing Evolution under Different Objectives}
\label{ssec:objective_analysis}

\paragraph{Search trajectories.}
The observed runs exhibit different patterns of refinement and exploration. In the illustrated accuracy-only trajectories, search progresses from simple strategies toward increasingly elaborate designs, adding operations to address remaining errors. As accuracy approaches a plateau, further gains come with substantial increases in serving cost. As shown in Figure~\ref{fig:lme_evolution}, introducing per-session note extraction at build time from step 13 increases cost approximately sixfold for additional search accuracy. Figure~\ref{fig:locomo_lineage}a illustrates the predominantly additive nature of this process, with successive revisions extending a dominant chain. The illustrated joint accuracy--cost trajectory maintains lower serving costs and exhibits a more branched lineage, as shown in Figure~\ref{fig:locomo_lineage}b, exploring alternative designs and removing costly operations. Lower-accuracy candidates can remain nondominated and serve as starting points for subsequent revisions.

\paragraph{Resulting memory designs.}
The selected programs allocate LLM computation differently across the memory lifecycle. \sysname[-A\textsubscript{LME}] extracts atomic notes from each session at build time, which accounts for about 92\% of its serving cost. \sysname[-A\textsubscript{LCM}] distills facts into speaker profiles and runs a heavier reader over them at query time. Both accuracy--cost designs instead retrieve raw conversation text and compute selectively at query time: \sysname[-P\textsubscript{LME}] uses BM25~\citep{robertson2009bm25} without build-time model calls, and \sysname[-P\textsubscript{LCM}] retrieves raw turns and consults extracted facts only for aggregation questions. These designs use 90\% and 40\% fewer serving calls than their counterparts on LongMemEval and LoCoMo, respectively. The larger reduction on LongMemEval is consistent with its workload: each history serves a single query, so memory construction costs cannot be shared across multiple queries. \sysname[-P\textsubscript{LME}] avoids build-time model calls entirely. These programs retain competitive accuracy with fewer serving calls. Despite their differences in computation allocation, the selected designs share the use of raw evidence and query-dependent composition, motivating the component analysis below.

\subsection{Emergent Memory Designs and Component Analysis}
\label{sec:analysis}

Across the two benchmarks and optimization objectives, the discovered memory systems exhibit two recurring design patterns: retaining raw conversational content and retrieving it as evidence, and routing queries to different composition strategies according to question type. To examine these patterns in detail, we conduct a case study and component analysis of \sysname[-A\textsubscript{LME}], a representative design that incorporates both.

Figure~\ref{fig:memorph_a} illustrates the workflow of \sysname[-A\textsubscript{LME}], highlighting two core designs. \textbf{(1) It organizes memory hierarchically while retaining both extracted and raw evidence.} During construction, an LLM extracts atomic notes~\citep{chen2024dense} from each session, while the original conversation is preserved as individual rounds; both are jointly indexed using dense and lexical retrieval. At retrieval time, item-level scores are aggregated to rank sessions, and the system returns session-level note bundles alongside up to five highest-ranked raw conversation rounds. \textbf{(2) Query-Routed Scaffolding (RQS) conditions composition on query type.} RQS routes each query by intent and runs an intent-specific scaffold that pairs LLM extraction with deterministic computation, for instance date arithmetic for temporal queries and per-session item extraction followed by programmatic counting for aggregation queries. A final deliberation call applies intent-specific instructions to produce a draft answer, which is passed alongside the evidence to the fixed QA model for final answer generation. Appendix~\ref{app:compose} details the routing and scaffolds.

\input{sections/tab_compose_ablation}
\input{sections/fig_compose_transfer}

\textbf{First, we examine how retaining and retrieving raw conversations affects overall performance.} We remove the retrieved raw rounds from the answer context while retaining them in the index and leaving the remaining pipeline unchanged. This reduces accuracy by an absolute 12.17\% on LoCoMo and 6.50\% on LongMemEval, as shown in Table~\ref{tab:compose_ablation}. The largest decline on LongMemEval occurs for single-session-assistant questions, where accuracy drops by an absolute 20.00\%. Category-level results are provided in Table~\ref{tab:ablation_lme} in Appendix~\ref{app:rawrounds}. These results highlight the importance of raw conversations for recovering details omitted from extracted memories. In contrast, all five evaluated human-designed memory systems discard the original conversations after extracting facts during construction or omit them from the evidence provided to the QA model. This can leave the QA model without fine-grained semantic information needed for subsequent questions. Our error attribution analysis identifies incomplete retrieved evidence as a major source of errors in these systems. Estimated evidence completeness is strongly positively associated with overall accuracy. Appendix~\ref{app:evidence} details the estimation procedure and an additional analysis of incorrect answers, with results shown in Figures~\ref{fig:evidence} and~\ref{fig:evidence_acc}.

\textbf{We next evaluate the effectiveness and transferability of RQS.} We compare RQS with default evidence rendering using the same recorded retrieval and fixed QA settings. Removing RQS from \sysname[-A\textsubscript{LME}] reduces accuracy by an absolute 4.44\% on LoCoMo and 6.25\% on LongMemEval, with the largest declines on temporal questions, as shown in Tables~\ref{tab:ablation_lme} and~\ref{tab:ablation_locomo}. This finding is consistent with the difficulties in temporal reasoning and counting identified by the proposer and diagnosis agents during evolution with both GPT-4.1 mini and Qwen backbones. RQS addresses these failure modes by combining LLM-based extraction with deterministic computation \citep{gao2023pal,chen2023pot}. The separation of composition from memory construction and retrieval further allows us to transplant RQS into five human-designed memory systems through a fixed input adapter. Across the five systems, RQS yields absolute accuracy gains of 2.75--9.25\% on LongMemEval under the paired evaluation protocol, as shown in Figure~\ref{fig:compose_transfer}. Mem0 shows the smallest improvement, at 2.75\%. Our error analysis suggests that its failures are dominated by missing retrieved information, whereas RQS primarily addresses cases in which sufficient evidence is available but the QA model fails to identify or use it effectively.

\input{sections/related_work}

\section{Conclusion}
\label{sec:conclusion}
We introduced \sysname{} to study agent memory design through whole-program evolution from minimal seeds. Across two benchmarks, the evolved programs instantiate two recurring design patterns: retaining raw conversations as retrievable evidence and processing retrieved evidence through intent-specific scaffolds before answering. Ablations highlight the value of preserving raw evidence, while transplanting the evolved composition module improves all five evaluated human-designed systems. Under the evaluated workloads, incorporating serving cost into the search objective favors selective query-time processing over LLM-based memory construction, while retaining competitive accuracy. Together, these findings illustrate how program evolution, combined with component analysis and transplantation, can produce effective memory systems and empirically grounded design lessons.

\section*{Limitations}

Our findings concern conversational memory on two QA benchmarks, with detailed component analysis centered on one representative program. Transfer and transplantation experiments extend the evaluation across models, benchmarks, and memory systems, while applicability to experiential memory and open-ended agents remains to be established. Evolutionary trajectories and resulting programs can vary across runs, and program selection and design interpretation involve empirical judgment. Our cost comparisons cover serving token charges for LLM and embedding calls, excluding search expenses, local computation, and latency. Their implications depend on workload, particularly how memory construction costs are amortized across queries.

\section*{AI Assistance Statement}

AI assistants were used to help revise manuscript text and edit figures. The authors made the final decisions on the content and presentation and take responsibility for the submitted work.

\FloatBarrier
\bibliography{custom}

\clearpage
\appendix

\input{sections/appendix_interface}

\input{sections/appendix_search}

\input{sections/appendix_data_protocol}

\input{sections/appendix_supp}

\input{sections/appendix_pending}

\end{document}

%% file: sections/abstract.tex
Program evolution provides a promising approach to discover agent memory designs and study the choices underlying their effectiveness. We introduce \sysname{}, a framework that evolves whole memory programs from minimal seeds through a standardized lifecycle interface and an iterative diagnosis--proposal loop. Experiments on LoCoMo and LongMemEval yield an absolute accuracy gain of up to 12\% over manually designed baselines at 35\% of the serving cost, with further evaluations supporting transfer across serving backbones and benchmarks. Empirical analyses and ablations identify and validate two recurring design patterns: retaining raw conversations as retrievable evidence preserves details omitted from extracted memories, while composing retrieved evidence according to query intent helps the answering model use it effectively. Transplanting the composition module into five existing memory systems improves accuracy across all five. Beyond these patterns, the search objective shapes where computation goes: optimizing serving cost alongside accuracy shifts computation from LLM-based memory construction to selective processing at query time while retaining competitive accuracy. Together, these findings offer empirical lessons for agent memory design. Code is available at \url{https://github.com/ash-ding/MAPE}.

%% file: sections/introduction.tex
\section{Introduction}
\label{sec:intro}

Agents powered by large language models, such as personal assistants and conversational companions~\citep{wang2024agentsurvey,xi2023rise}, increasingly depend on persistent memory of user information to maintain continuity across conversations. Agent memory systems~\cite{packer2023memgpt,chhikara2025mem0,xu2025amem,rasmussen2025zep,liu2026simplemem} address this need by transforming interaction history into memory units, organizing them in a persistent store, and retrieving relevant information to support subsequent responses. Prior work has advanced memory representation, organization, maintenance, and retrieval~\citep{zhang2024memsurvey}, yet these components are often designed and refined through human expertise.

Meanwhile, LLM-driven evolutionary search has emerged as a tool for automated scientific discovery and algorithm design~\citep{romeraparedes2024funsearch,liu2024eoh,ma2024eureka}. Systems such as AlphaEvolve~\citep{novikov2025alphaevolve} employ an LLM to propose program modifications and automated evaluation to guide subsequent proposals, yielding mathematical constructions and algorithms that outperform established designs. Recent work extends this paradigm to agent memory, evolving prompts, configurations, or executable pipelines to improve memory performance~(\citealp{deng2026memtrace}; \citealp{liu2026evolvemem}; \citealp{liu2026mempro}).

These approaches often constrain the search space to selected components or start from established memory architectures, and primarily emphasize accuracy. We explore a complementary direction: evolving entire memory programs from minimal seeds to study the designs that emerge under different optimization objectives. Beyond finding better programs, we seek to understand how evolution implements effective memory design choices, whether their benefits extend to other memory systems, and how different optimization objectives shape those choices. Explicitly considering serving cost also allows us to study how evolution allocates computation across the memory lifecycle. Specifically, we ask: \circlednum{1} \textit{What effective memory system designs can whole-program evolution discover from minimal seeds?} \circlednum{2} \textit{How do different optimization objectives shape the evolutionary process and the resulting designs?}

To investigate these questions, we develop \sysname{}, a framework that evolves agent memory systems through whole-program evolution, as illustrated in Figure~\ref{fig:overview}. A standardized lifecycle interface exposes memory construction, retrieval, and composition~\citep{xu2026com} as programmable stages, so that the entire pipeline is open to revision under a common evaluation protocol. An iterative diagnosis--proposal loop uses execution traces to identify failure patterns and guide subsequent revisions~\citep{madaan2023selfrefine,shinn2023reflexion}. Search starts from minimal seeds, such as a program that stores no memory. The framework tracks serving cost across the memory lifecycle and supports both accuracy-only and joint accuracy--cost search.

We run \sysname{} on LoCoMo~\citep{maharana2024locomo} and LongMemEval~\citep{wu2025longmemeval} under both objectives, yielding four experimental settings. For each, we analyze the evolutionary trajectory and the resulting memory system. All selected systems outperform the evaluated human-designed baselines on their respective benchmarks. On LoCoMo, we observe an absolute accuracy gain of up to 12\% at 35\% of the serving cost. Further evaluations include cross-backbone validation using GPT-4.1 mini and Qwen3-30B-A3B-Instruct-2507~\citep{yang2025qwen3}, as well as bidirectional cross-benchmark transfer between LoCoMo and LongMemEval, all without additional evolution. These results support the transferability of the discovered memory designs across serving backbones and benchmarks.

Across the discovered systems, we identify two recurring design patterns: retaining raw conversations as retrievable evidence and processing retrieved evidence through intent-specific scaffolds before answering. A case study and component ablations examine the contribution of each pattern in a representative system. Retaining raw evidence helps preserve details omitted from extracted memories, while error attribution across human-designed systems identifies incomplete retrieved information as a major bottleneck. The composition module routes queries to intent-specific scaffolds that combine targeted extraction, deterministic computation, and deliberation to support the answering model. Transplantation experiments test this composition strategy across five human-designed memory systems, improving accuracy in every case and demonstrating its utility beyond the evolved program.

Comparing the two objective settings reveals differences in both search behavior and computation allocation. In the observed runs, accuracy-only search progressively adds operations, accepting substantial cost increases for marginal accuracy gains. The illustrated joint accuracy--cost trajectory maintains lower serving costs and explores more branched candidate lineages, retaining inexpensive alternatives as starting points for further evolution. The resulting designs differ in where they spend computation. The selected accuracy-only programs invest in LLM-based memory construction, whereas the selected accuracy--cost programs reduce reliance on it in favor of selective computation at query time, using 90\% and 40\% fewer serving calls on LongMemEval and LoCoMo while retaining strong evaluation-set accuracy.

Our contributions are threefold:
\begin{itemize}
    \item We introduce \sysname{}, a framework for discovering and studying agent memory designs through whole-program evolution from minimal seeds under a standardized lifecycle interface.
    \item We identify raw-conversation retention and query-dependent composition as recurring design patterns, validate their contributions through ablations, and demonstrate the transferability of composition by transplanting it into five human-designed systems.
    \item We show how optimization objectives shape evolutionary trajectories and computation allocation, with joint accuracy--cost search shifting computation from LLM-based memory construction to selective processing at query time.
\end{itemize}

%% file: sections/method.tex
\section{The \texorpdfstring{\sysname{}}{MAPE} Framework}
\label{sec:method}

\input{sections/fig_overview}

\sysname{} provides a framework for studying agent memory design through whole-program evolution. This section defines the memory task, program search space, and optimization objectives, then describes the evolutionary process and final program selection.

\subsection{Problem Formulation}
\label{ssec:problem}

We consider a conversation history $H_i$ and its associated query set $\mathcal Q_{H_i}=\{q_{i,1},\ldots,q_{i,n_i}\}$. A memory system $m$ organizes information from $H_i$ to support history-grounded question answering through four stages: \textit{build} ($B_m$), \textit{retrieve} ($R_m$), \textit{compose} ($K_m$), and \textit{answer} ($G_\theta$). Memory construction is performed once for each history, followed by retrieval, composition, and answering for each associated query.

\paragraph{Build.}
$B_m$ maps the conversation history $H_i$ to a persistent memory state $M_i=B_m(H_i)$. This state comprises retained content and the data structures used to organize and index it.

\paragraph{Retrieve.}
$R_m$ selects evidence from $M_i$ for query $q_{i,j}$, yielding $E_{i,j}=R_m(M_i,q_{i,j})$. The retrieved evidence provides query-relevant information.

\paragraph{Compose.}
$K_m$ transforms query $q_{i,j}$ and retrieved evidence $E_{i,j}$ into answer context $c_{i,j}=K_m(q_{i,j},E_{i,j})$. Inspired by CoM~\citep{xu2026com}, we introduce this additional stage to explicitly separate post-retrieval evidence processing from retrieval itself.

\paragraph{Answer.}
$G_\theta$ maps query $q_{i,j}$ and composed context $c_{i,j}$ to a predicted answer $\hat y_{i,j}=G_\theta(q_{i,j},c_{i,j})$. Here, $\theta$ specifies the answering model and prompt.

\subsection{Program Search Space}
\label{ssec:contract}

We construct a program search space for exploring diverse memory system designs under a common lifecycle interface $\mathcal I$. Candidate programs implement memory construction and retrieval and can additionally define how retrieved evidence is composed into the answer context. Evolution can modify these stage implementations and their supporting code. This allows evolution to explore alternative representations, retrieval mechanisms, and evidence-processing procedures, including coordinated changes across stages. Appendix~\ref{app:interface} provides a simplified Python class interface.

The resulting search space $\mathcal M_{\mathcal I}$ is bounded by interface compatibility and a fixed answering procedure. Within each evaluation setting, the answering model, prompt, and evaluation protocol are held constant, enabling comparisons among the discovered memory designs under shared evaluation conditions.

\subsection{Search Objectives}
\label{ssec:objectives}
\label{ssec:cost}

Throughout an evolutionary run, each candidate program $m$ is evaluated on a fixed search set $\mathcal D$. Serving LLM and embedding calls use the same designated LLM and embedding model, respectively, across all lifecycle stages. This shared configuration enables controlled comparisons of candidate programs in terms of answer accuracy and lifecycle serving cost.

\paragraph{Accuracy.}
For each query $q_{i,j}$, we use an LLM judge~\citep{zheng2023judging} to evaluate answer correctness:
\begin{equation}
z_{i,j}=J_{\mathrm{LLM}}(r,q_{i,j},\hat y_{i,j},y_{i,j})\in\{0,1\},
\label{eq:judge}
\end{equation}
where $r$ is the benchmark-specific rubric prompt, $\hat y_{i,j}$ the predicted answer, and $y_{i,j}$ the ground truth. For the search objective, we first average correctness judgments within each history and then take an unweighted mean across histories to obtain $A_{\mathcal D}(m)$. Reported benchmark accuracies instead pool correctness judgments across all evaluated queries, as described in Appendix~\ref{app:models_evaluation}.

\paragraph{Lifecycle serving cost.}
We measure the token-based cost of constructing memory for each history and serving its associated queries through retrieval, composition, and answering. For each history $H_i$ and lifecycle stage $s$, we record the total LLM input, LLM output, and embedding tokens, denoted by $T_{i,s}^{\mathrm{in}}$, $T_{i,s}^{\mathrm{out}}$, and $T_{i,s}^{\mathrm{emb}}$. We compute the stage cost using the corresponding per-token prices:
\begin{equation}
C_{i,s}(m)=p^{\mathrm{in}}T_{i,s}^{\mathrm{in}}+p^{\mathrm{out}}T_{i,s}^{\mathrm{out}}+p^{\mathrm{emb}}T_{i,s}^{\mathrm{emb}},
\label{eq:cost}
\end{equation}
We sum the costs across all four stages and divide by the number of answered queries for each history, amortizing memory construction over the queries it serves. We then take an unweighted mean across histories and scale the result to dollars per 1{,}000 queries, yielding $C_{\mathcal D}(m)$. This metric includes LLM and embedding calls but excludes judge and search-agent calls, local computation, and storage. It does not measure latency. Pricing details are provided in Appendix~\ref{app:data_protocol}.

With both metrics recorded, we run the evolutionary search in two modes: \sysname[-A] optimizes accuracy alone, whereas \sysname[-P] treats accuracy and lifecycle serving cost as separate objectives. Its Pareto frontier contains programs for which no other evaluated candidate achieves at least equal accuracy and no greater cost, with a strict improvement in at least one objective~\citep{deb2002nsga2}. The proposer uses this frontier to guide subsequent proposals, while the archive retains all evaluated candidates.

\subsection{Evolutionary Search Loop}
\label{ssec:loop}

\sysname{} begins by evaluating the seed programs and placing them in an archive. At each iteration, the proposer consults archived implementations, evaluation results, and diagnoses, selects parent programs, and produces a new candidate with a design hypothesis. The hypothesis states the intended change and its expected benefit, providing a basis for subsequent diagnosis.

The candidate first undergoes a small sanity evaluation to identify execution errors. The proposer can repair these errors within a fixed budget, and candidates that pass are evaluated on the full fixed search set to obtain accuracy and lifecycle serving cost. A separate diagnosis agent then examines the implementation, design hypothesis, evaluation results, and available execution traces to assess the observed outcome and identify potential causes of failure. The diagnosis informs subsequent proposals without modifying the evaluated candidate.

After evaluation and diagnosis, the candidate implementation, parent references, hypothesis, results, traces, and diagnosis are added to the archive. All evaluated candidates are retained, including dominated designs, allowing later proposals to revisit earlier implementations rather than extend only the current best program. Search continues until the proposal budget is exhausted. The evaluation set remains unavailable throughout evolution and is used only afterward in the program-selection procedure described below. Figure~\ref{fig:overview} illustrates the loop, and Appendix~\ref{app:evolution_setup} provides the initialization, search budgets, and execution settings.

\subsection{Program Selection}
\label{ssec:selection}

After evolution, we shortlist candidates using search-set accuracy for \sysname[-A] and the search-set accuracy--cost Pareto frontier for \sysname[-P], with additional inclusion criteria described in Appendix~\ref{app:program_selection}. The shortlisted programs are frozen and evaluated on a larger evaluation set that remains inaccessible during evolution. We then inspect their implementations and exclude programs with impractical scaling behavior, such as exhaustive LLM processing of the history during retrieval. For each setting, we report the remaining program with the highest evaluation-set accuracy. The selected programs are denoted \sysname[-A\textsubscript{LCM}], \sysname[-P\textsubscript{LCM}], \sysname[-A\textsubscript{LME}], and \sysname[-P\textsubscript{LME}], where \mono{A} and \mono{P} indicate the accuracy-only and joint accuracy--cost search objectives, and \mono{LCM} and \mono{LME} denote evolution on LoCoMo and LongMemEval, respectively.

%% file: sections/fig_overview.tex
\begin{figure*}[t]
\centering
\includegraphics[width=\textwidth]{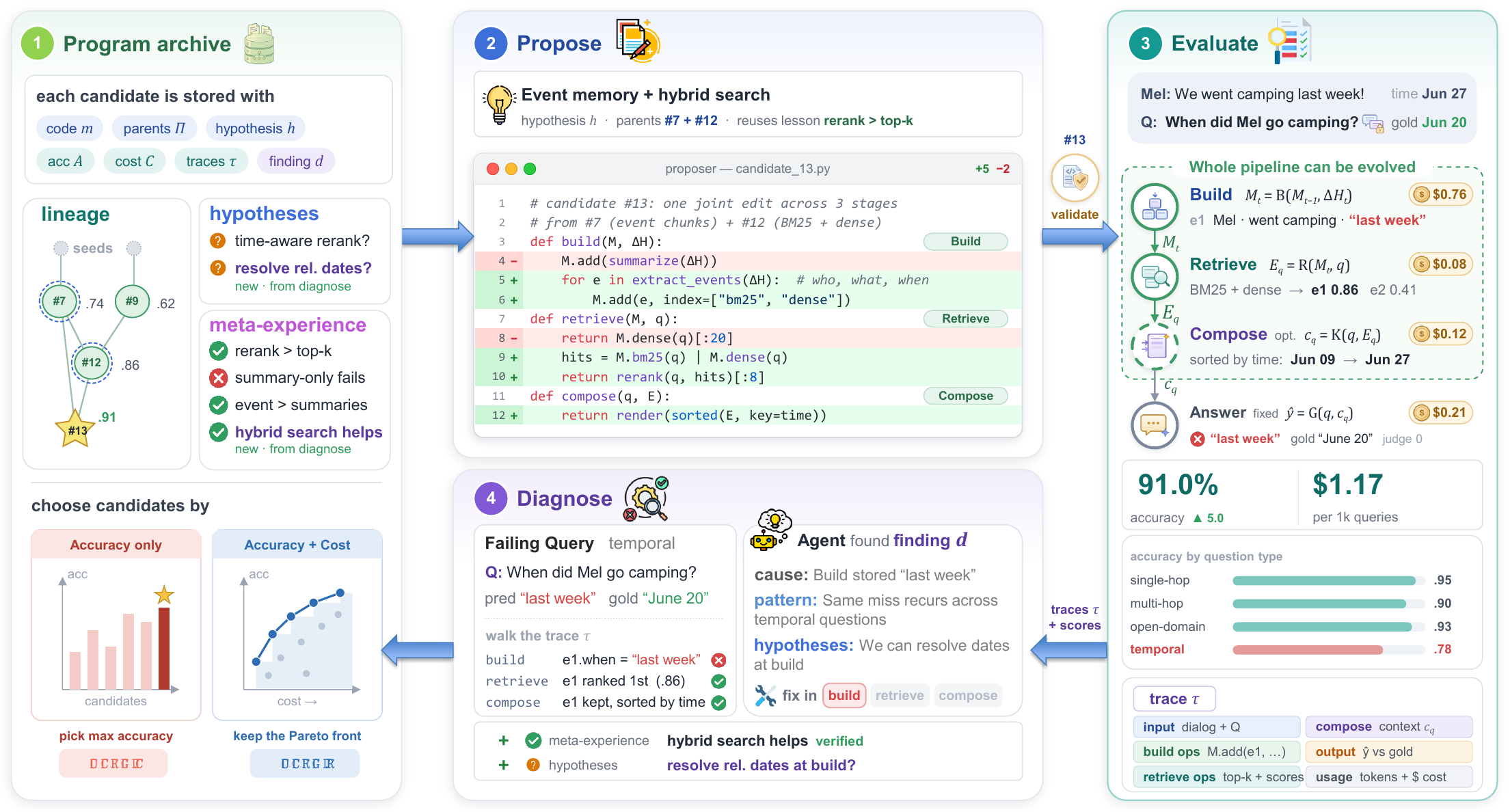}
\caption{Overview of \sysname{}'s program evolution framework. Candidate programs are proposed using archived designs and diagnostic feedback, checked for execution errors, and evaluated on the search set. Their implementations, results, and diagnoses are retained to guide subsequent proposals. Evolution follows either an accuracy-only objective (\sysname[-A]) or joint accuracy--cost objectives (\sysname[-P]). Shortlisted programs undergo held-out evaluation for final selection. }
\label{fig:overview}
\end{figure*}

%% file: sections/tab_main.tex
\input{sections/tab_locomo}
\input{sections/tab_longmemeval}

%% file: sections/tab_locomo.tex
\begin{table*}[!t]
\centering
\small
\setlength{\tabcolsep}{5pt}
\begin{tabularx}{\textwidth}{l >{\centering\arraybackslash}X >{\centering\arraybackslash}X >{\centering\arraybackslash}X >{\centering\arraybackslash}X >{\centering\arraybackslash}X >{\centering\arraybackslash}X}
\toprule
\textbf{Method} & \textbf{Multi} & \textbf{Open} & \textbf{Single} & \textbf{Temp} & \textbf{Avg} & \textbf{Cost} (\$/1k)$\downarrow$ \\
\midrule
\rowcolor{black!6}\multicolumn{7}{l}{\emph{GPT-4.1 mini}} \\
\midrule
Full Context & 71.55 & \underline{62.65} & 91.06 & 66.67 & 80.87 & 12.54 \\
RAG & 33.47 & 39.76 & 59.70 & 47.29 & 51.19 & 0.25 \\
\midrule
Mem0 & 62.34 & 51.81 & 85.01 & 70.93 & 75.98 & 1.10 \\
A-Mem & 50.63 & 54.22 & 75.38 & 63.95 & 67.25 & 8.60 \\
LightMem & 59.00 & 42.17 & 68.09 & 52.71 & 61.74 & 0.56 \\
SimpleMem & 69.46 & 57.83 & 90.37 & \underline{71.71} & 80.80 & 5.46 \\
Zep & 54.81 & 43.37 & 63.41 & 50.39 & 58.00 & 11.76 \\
\midrule
\sysname[-A\textsubscript{LCM}] & \textbf{83.68} & \textbf{65.06} & \textbf{94.91} & \textbf{79.84} & \textbf{87.94} & 7.21 \\
\sysname[-P\textsubscript{LCM}] & \underline{79.49} & 59.03 & \underline{91.47} & \textbf{79.84} & \underline{85.18} & 2.03 \\
\midrule
\rowcolor{black!6}\multicolumn{7}{l}{\emph{Qwen3-30B-A3B-Instruct-2507}} \\
\midrule
Full Context & 58.16 & 53.01 & 89.41 & 31.78 & 70.01 & 7.07 \\
RAG & 27.62 & 38.55 & 58.05 & 16.67 & 43.08 & 0.14 \\
\midrule
Mem0 & 60.25 & 56.63 & 77.44 & 41.86 & 65.95 & 0.61 \\
A-Mem & 44.77 & 37.35 & 67.95 & 24.42 & 53.18 & 5.71 \\
LightMem & 54.81 & 45.78 & 76.20 & 28.68 & 60.98 & 0.48 \\
SimpleMem & 61.09 & 46.99 & 86.52 & 35.66 & 69.32 & 3.24 \\
Zep & 50.21 & 44.58 & 57.22 & 30.62 & 49.89 & 6.31 \\
\midrule
\sysname[-A\textsubscript{LCM}] & \textbf{78.66} & \textbf{68.67} & \textbf{92.02} & \underline{43.80} & \underline{78.58} & 3.47 \\
\sysname[-P\textsubscript{LCM}] & \underline{72.80} & \underline{65.06} & \underline{89.68} & \textbf{70.93} & \textbf{81.33} & 1.14 \\
\bottomrule
\end{tabularx}
\caption{Accuracy and lifecycle serving cost on the LoCoMo evaluation set with two serving backbones. Multi, Open, Single, and Temp denote multi-hop, open-domain, single-hop, and temporal questions, respectively; Avg denotes accuracy over all questions. Accuracy is reported in \% and cost in dollars per 1,000 queries. Within each backbone, the highest and second-highest accuracies per column are \textbf{bold} and \underline{underlined}, respectively.}
\label{tab:locomo}
\end{table*}

%% file: sections/tab_longmemeval.tex
\begin{table*}[!t]
\centering
\small
\setlength{\tabcolsep}{5pt}
\begin{tabularx}{\textwidth}{l >{\centering\arraybackslash}X >{\centering\arraybackslash}X >{\centering\arraybackslash}X >{\centering\arraybackslash}X >{\centering\arraybackslash}X >{\centering\arraybackslash}X >{\centering\arraybackslash}X c}
\toprule
\textbf{Method} & \textbf{Temp} & \textbf{Multi} & \textbf{Know} & \textbf{User} & \textbf{Asst} & \textbf{Pref} & \textbf{Avg} & \textbf{Cost} (\$/1k)$\downarrow$ \\
\midrule
\rowcolor{black!6}\multicolumn{9}{l}{\emph{Qwen3-30B-A3B-Instruct-2507}} \\
\midrule
Full Context & 43.40 & 41.51 & 74.60 & 91.07 & 93.33 & 37.50 & 59.75 & 22.27 \\
RAG & 43.40 & 53.77 & \underline{85.71} & \underline{94.64} & \underline{97.78} & 41.67 & 66.00 & 2.62 \\
\midrule
Mem0 & 31.13 & 38.68 & 60.32 & 80.36 & 48.89 & 41.67 & 47.25 & 93.55 \\
A-Mem & 42.45 & 45.28 & 60.32 & 82.14 & \textbf{100.00} & 41.67 & 58.00 & 661.97 \\
LightMem & 49.06 & \underline{57.55} & 79.37 & \textbf{96.43} & 26.67 & 37.50 & 59.50 & 26.59 \\
SimpleMem & 40.57 & 34.91 & 55.56 & 73.21 & 82.22 & 33.33 & 50.25 & 144.74 \\
Zep & 37.74 & 30.19 & 58.73 & 80.36 & 60.00 & 33.33 & 47.25 & 2367.28 \\
\midrule
\sysname[-A\textsubscript{LME}] & \textbf{80.19} & \textbf{58.49} & \textbf{88.89} & 89.29 & 93.33 & \textbf{54.17} & \textbf{77.00} & 52.84 \\
\sysname[-P\textsubscript{LME}] & \underline{70.75} & \underline{57.55} & 84.13 & 85.71 & 91.11 & \underline{45.83} & \underline{72.25} & 5.76 \\
\bottomrule
\end{tabularx}
\caption{Accuracy and lifecycle serving cost on the LongMemEval evaluation set with Qwen3-30B-A3B-Instruct-2507 as the serving backbone. Temp, Multi, Know, User, Asst, and Pref denote temporal reasoning, multi-session, knowledge-update, single-session-user, single-session-assistant, and single-session-preference questions, respectively; Avg denotes accuracy over all questions. Accuracy is reported in \% and cost in dollars per 1,000 queries. The highest and second-highest accuracies in each column are \textbf{bold} and \underline{underlined}, respectively.}
\label{tab:longmemeval}
\end{table*}

%% file: sections/experimental_setup.tex
\section{Experimental Setup}
\label{sec:setup}

\paragraph{Benchmarks and protocol.}
We evaluate on LoCoMo~\citep{maharana2024locomo} and LongMemEval-S~\citep{wu2025longmemeval}, using fixed 20\% search and 80\% evaluation partitions. LoCoMo is split by conversation, while LongMemEval-S is split by question instance together with its associated history. During evolution, candidates are evaluated exclusively on the search partition. The evaluation partition is used afterward for program selection and performance reporting. Appendix~\ref{app:split_construction} provides the split construction details.

\paragraph{Baselines.}
We compare against Full Context, which supplies conversation history directly to the QA model, and RAG~\citep{lewis2020rag}, which retrieves messages by embedding similarity. We also evaluate five human-designed memory systems: Mem0~\citep{chhikara2025mem0}, A-Mem~\citep{xu2025amem}, LightMem~\citep{fang2025lightmem}, SimpleMem~\citep{liu2026simplemem}, and Zep~\citep{rasmussen2025zep}. Lightweight adapters connect their implementations to the shared evaluation interface without redesigning their memory methods. Appendix~\ref{app:baselines} describes the configurations and integration adjustments.

\paragraph{Models and metrics.}
We report LoCoMo results with GPT-4.1 mini~\citep{openai2025gpt41} and Qwen3-30B-A3B-Instruct-2507~\citep{yang2025qwen3}, and LongMemEval results with Qwen. Cross-benchmark transfer is also evaluated with Qwen. Within each comparison, the serving LLM and QA settings are shared across systems, with text-embedding-3-small as the designated embedding model. GPT-5 mini evaluates answer correctness using benchmark-specific rubrics~\citep{zheng2023judging}. We report accuracy pooled across evaluated queries and lifecycle serving cost in dollars per 1,000 queries. Section~\ref{ssec:objectives} defines the cost aggregation, and Appendix~\ref{app:models_evaluation} provides the reporting protocol and model, generation, and pricing details.

%% file: sections/fig_memorph_a.tex
\begin{figure*}[t]
\centering
\includegraphics[width=0.9\textwidth]{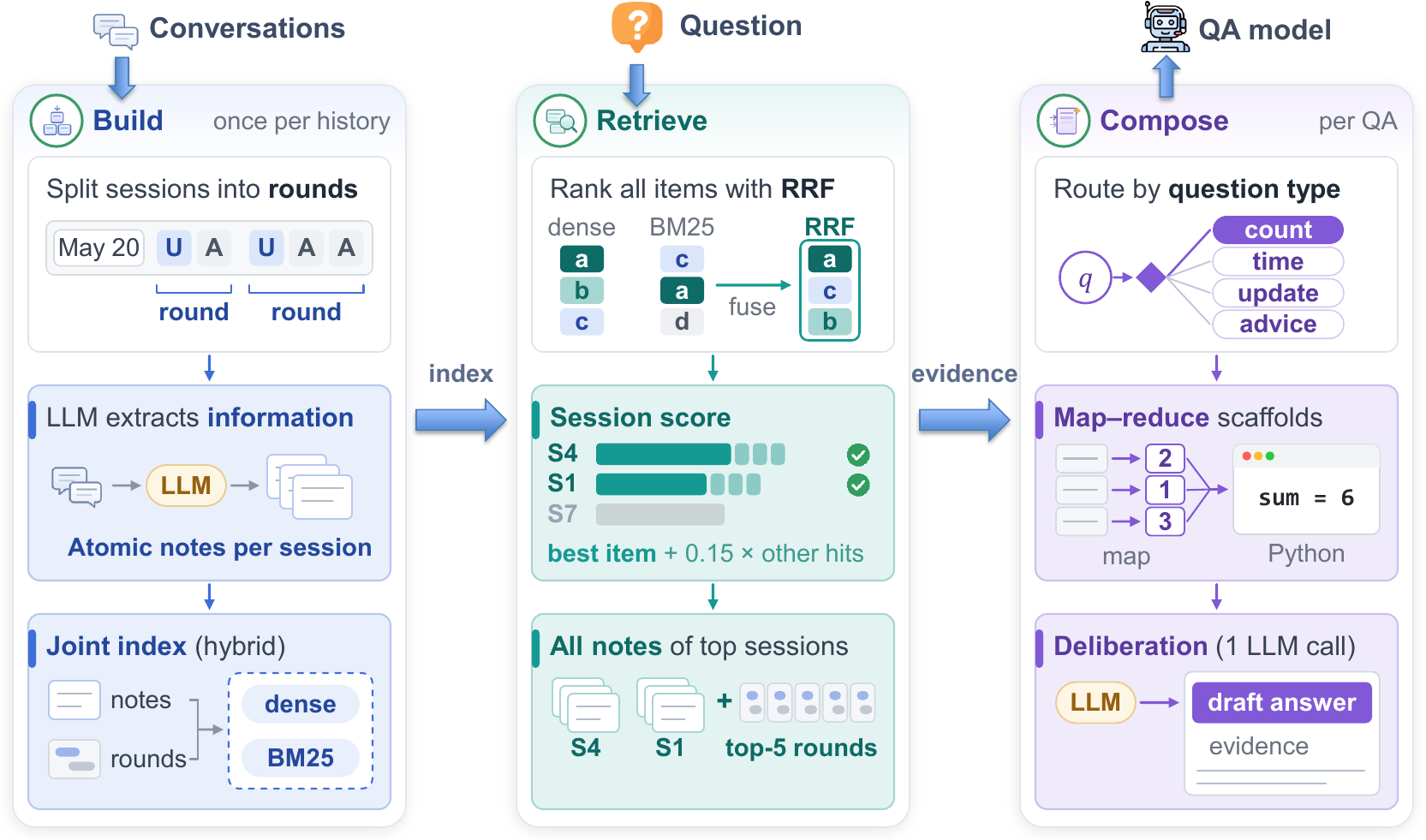}
\caption{Pipeline of \sysname[-A\textsubscript{LME}], evolved under the accuracy-only objective on LongMemEval. \emph{Build} retains both extracted atomic notes and raw conversations. \emph{Retrieve} combines dense and sparse rankings through reciprocal rank fusion (RRF) and returns relevant notes and raw conversational evidence. \emph{Compose} implements Query-Routed Scaffolding (RQS), which routes queries to type-specific scaffolds according to query intent, preparing context for the answering model.}
\label{fig:memorph_a}
\end{figure*}

%% file: sections/tab_compose_ablation.tex
\begin{table}[t]
\centering
\small
\setlength{\tabcolsep}{4pt}
\begin{tabular}{@{}lcc@{}}
\toprule
\multirow{2}{*}{\textbf{Method}} & \multicolumn{2}{c}{\textbf{Overall Accuracy}} \\
\cmidrule(l){2-3}
 & \textbf{LoCoMo} & \textbf{LongMemEval} \\
\midrule
\sysname[-A\textsubscript{LME}] & \textbf{75.90} & \textbf{77.00} \\
\quad w/o raw rounds & 63.73 & 70.50 \\
\quad w/o RQS & 71.46 & 70.75 \\
\quad w/o both & 63.12 & 63.00 \\
\bottomrule
\end{tabular}
\caption{Ablation of \sysname[-A\textsubscript{LME}] on the LoCoMo and LongMemEval evaluation sets using the Qwen3-30B-A3B-Instruct-2507 serving backbone. \emph{W/o raw rounds} retains only retrieved notes; \emph{w/o RQS} bypasses Query-Routed Scaffolding (RQS); \emph{w/o both} applies both ablations. Accuracy is reported in \%.}
\label{tab:compose_ablation}
\end{table}

%% file: sections/fig_compose_transfer.tex
\begin{figure}[t]
\centering
\includegraphics[width=\columnwidth]{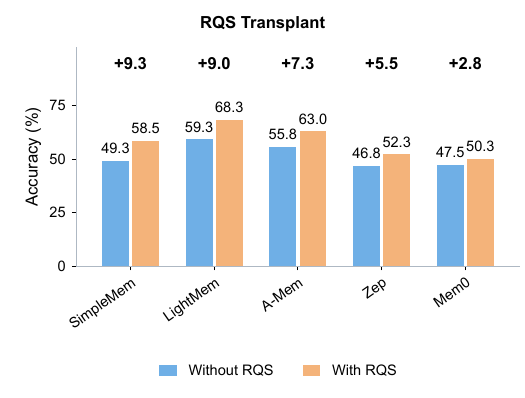}
\caption{RQS transplanted from \sysname[-A\textsubscript{LME}] into five human-designed memory systems, evaluated on the LongMemEval evaluation set with Qwen3-30B-A3B-Instruct-2507 as the serving backbone. Bold values indicate absolute accuracy gains in \%.}
\label{fig:compose_transfer}
\end{figure}

%% file: sections/related_work.tex
\section{Related Work}
\label{sec:related}

\paragraph{Agent memory design and empirical analysis.}
Agent memory systems organize interaction histories through reflection, selective updates, and hierarchical storage, as exemplified by Generative Agents, MemoryBank, and MemGPT \citep{park2023generative,zhong2024memorybank,packer2023memgpt}. Mem0 extracts and consolidates memories \citep{chhikara2025mem0}, while Zep and A-Mem organize them through temporal knowledge graphs and linked networks, respectively \citep{rasmussen2025zep,xu2025amem}. LightMem combines topic-based organization with offline consolidation \citep{fang2025lightmem}. SimpleMem combines semantic compression with intent-aware retrieval \citep{liu2026simplemem}. Mem0, Zep, A-Mem, LightMem, and SimpleMem serve as baselines in our empirical analysis.

Prior work has explored preserving conversational histories and processing evidence after retrieval. CoM combines lightweight memory construction with query-oriented organization of retrieved fragments into inference chains \citep{xu2026com}. CoM also informs our inclusion of a programmable composition stage. DeferMem and LazyMem shift evidence processing to query time, using learned models to produce relevant evidence from retrieved histories \citep{yin2026defermem,yu2026lazymem}. ReFind instead enables an agent to search raw conversation logs directly \citep{li2026refind}. These approaches differ in whether they primarily improve access to evidence or transform retrieved evidence for answering.

Recent empirical studies further examine the value of memory construction relative to raw-history retrieval. Raw-RAG demonstrates the competitiveness of dense retrieval and reranking over raw histories \citep{bao2026rawrag}. Memory Tax distinguishes the benefits of memory transformation from losses caused by removing or obscuring useful evidence \citep{he2026memorytax}. These findings provide complementary evidence for the importance of preserving and accessing source information. \sysname{} examines memory design choices through program evolution and evaluates the resulting components through ablation and transplantation. Its query-routed scaffolding combines intent-specific evidence processing with deterministic computation and deliberation.

\paragraph{Evolution of agent memory systems.}
MetaMem evolves knowledge-utilization guidance \citep{xin2026metamem}. MemTrace uses attribution of failures across memory operations to guide prompt optimization \citep{deng2026memtrace}, while EvolveMem adapts retrieval and answer-generation configurations \citep{liu2026evolvemem}. MemEvolve jointly evolves experiential knowledge and memory architectures within a modular design space~\citep{zhang2025memevolve}. At the level of agent capabilities, Mem$^2$Evolve couples experience distillation with the creation of reusable tools and expert agents~\citep{cheng2026mem2evolve}. MemPro extends this direction to system-level evolution, treating memory construction--retrieval pipelines as executable programs subject to failure-guided revision \citep{liu2026mempro}. MemScope uses execution traces and runtime memory state to guide localized code repairs to an existing memory architecture~\citep{su2026memscope}. These studies span the evolution of stored experience, reusable capabilities, and memory system code. \sysname{} starts from minimal seeds and permits coordinated changes across memory construction, retrieval, and composition. By comparing accuracy-only and joint accuracy--cost objectives, we examine how search objectives shape memory designs and the allocation of computation across stages. Beyond evaluating the resulting programs, we use component ablations and transplantation across existing memory systems to assess which design choices contribute to their effectiveness and transfer beyond the evolved implementation.

%% file: sections/appendix_interface.tex
\section{Memory Program Interface}
\label{app:interface}

The \hyperref[code:memory_interface]{\texttt{MemoClass} interface}
provides a unified framework for implementing and evaluating memory systems.
It defines two required methods for memory construction and retrieval,
along with an optional method for composing retrieved memories into context.
Baseline implementations are connected to this interface through adapters,
enabling consistent evaluation alongside evolved memory systems.
Integration details are provided in Appendix~\ref{app:baselines}.

\begin{codebox}[label={code:memory_interface}]{Memory System Interface}
from abc import ABC, abstractmethod
from typing import Dict, List, Union
from common.recorder import Basic_Recorder

class MemoClass(ABC):
    @abstractmethod
    async def build_memory_from_data(
        self, recorder: Basic_Recorder
    ) -> None:
        ...

    @abstractmethod
    async def retrieve_memory_for_query(
        self, recorder: Basic_Recorder
    ) -> Dict:
        ...

    async def compose_memory_context(
        self,
        recorder: Basic_Recorder,
        retrieved: Dict,
    ) -> Union[str, List[str], None]:
        return None
\end{codebox}

%% file: sections/appendix_search.tex
\section{Evolutionary Search Details}
\label{app:search_details}

\subsection{Search Configuration}
\label{app:evolution_setup}
\paragraph{Initialization.}
\sysname[-P] is initialized with no-memory and full-context seeds, representing minimal memory overhead and unrestricted access to conversation history, respectively. These seeds provide contrasting starting points for exploring the trade-off between accuracy and cost. \sysname[-A] starts only from the no-memory seed to reduce architectural anchoring on full-context inference and encourage exploration beyond incremental revisions to that design.

\paragraph{Search budget.}
We perform three evolutionary runs per setting on LoCoMo and one per setting on LongMemEval. Each run consists of 20 steps, with one candidate program proposed at each step.

\paragraph{Execution and repair.}
Before full evaluation, each proposed candidate undergoes a sanity check using three questions from one search conversation on LoCoMo or two search instances on LongMemEval. If the check fails, the proposer receives the error trace and can repair the program up to twice, with a new check after each repair. Candidates that still fail are not evaluated on the full search set. Candidates that pass are evaluated on the entire search partition without progressive filtering or resampling between steps. Each proposer invocation allows up to 80 agent turns and 45 minutes, while each diagnosis invocation allows up to 40 turns and 30 minutes.

\paragraph{Search objective.}
\sysname[-P] jointly optimizes QA accuracy and lifecycle serving cost as two separate objectives. The proposer uses the accuracy--cost Pareto frontier to guide candidate generation. \sysname[-A] optimizes QA accuracy alone.

\paragraph{Search models.}
Both the proposer and diagnosis agents use Claude Opus 4.8 with reasoning effort set to \texttt{xhigh} throughout the evolutionary loop. During search, GPT-4.1 mini serves as the backbone for both LLM-based memory operations and QA generation on LoCoMo, while Qwen3-30B-A3B-Instruct-2507 serves these roles on LongMemEval. We use text-embedding-3-small for all embedding calls across both benchmarks.

\subsection{Program Selection}
\label{app:program_selection}
\paragraph{Shortlisting.}
For \sysname[-A], we shortlist the top 20\% of candidates ranked by search-set accuracy. For \sysname[-P], we shortlist all candidates on the search-set accuracy--cost Pareto frontier. Each shortlist additionally includes the program with the most direct children in that run's lineage, if it is not already included. On LoCoMo, we pool the shortlists from the three runs within each setting before final selection. Shortlisted programs are frozen and evaluated on the larger evaluation set, which remains unavailable throughout evolution.

\paragraph{Implementation screening and final selection.}
After evaluation, we manually inspect the shortlisted implementations and exclude programs with impractical scaling behavior, such as reliance on full-history inference or exhaustive LLM-based processing of the history during retrieval. For each setting, we report the remaining program with the highest evaluation-set accuracy. The evaluation set is thus used for final selection, but provides no feedback for program evolution.

%% file: sections/appendix_data_protocol.tex
\section{Experimental Setup Details}
\label{app:data_protocol}

\subsection{Data Splits}
\label{app:split_construction}
We use fixed, disjoint search and evaluation partitions. The evaluation partition remains unavailable during evolution and is subsequently used for final program selection and performance reporting. The 20/80 allocation applies to conversations in LoCoMo and question instances in LongMemEval.

\paragraph{LoCoMo.}
The dataset contains ten conversations. Following the implementation's fixed file order, we use the first two, \texttt{conv-26} and \texttt{conv-30}, for search and the remaining eight for evaluation.  Adversarial questions are excluded before evaluation. This yields 233 search questions and 1,307 evaluation questions. Table~\ref{tab:locomo_split} reports the resulting counts, including the excluded questions for completeness.

\begin{table}[!t]
\centering
\small
\setlength{\tabcolsep}{4pt}
\begin{tabularx}{\columnwidth}{Xrrr}
\toprule
\textbf{Category} & \textbf{Search} & \textbf{Evaluation} & \textbf{Total} \\
\midrule
Multi-hop & 43 & 239 & 282 \\
Temporal & 63 & 258 & 321 \\
Open-domain & 13 & 83 & 96 \\
Single-hop & 114 & 727 & 841 \\
\midrule
Evaluated questions & 233 & 1,307 & 1,540 \\
Adversarial (excluded) & 71 & 375 & 446 \\
\midrule
All questions & 304 & 1,682 & 1,986 \\
Conversations & 2 & 8 & 10 \\
\bottomrule
\end{tabularx}
\caption{LoCoMo partition sizes. Questions inherit the partition of their conversation; adversarial questions are excluded from evaluation.}
\label{tab:locomo_split}
\end{table}

\paragraph{LongMemEval.}

We use LongMemEval-S, which contains 500 question instances with associated conversation histories. We allocate 100 instances for search and 400 for evaluation, using stratified sampling across the six question types to maintain an approximately 20/80 split within each type. Within each question type, we sort question IDs and shuffle them with a shared Python random generator initialized with the fixed seed \texttt{longmemeval.search.v1}, processing types in sorted order. Search counts are allocated proportionally, with remaining slots assigned to the largest fractional remainders and ties resolved by type name. The remaining instances form the evaluation partition. Table~\ref{tab:longmemeval_split} reports the resulting distribution.

\begin{table}[!t]
\centering
\small
\setlength{\tabcolsep}{3pt}
\begin{tabularx}{\columnwidth}{Xrrr}
\toprule
\textbf{Question type} & \textbf{Search} & \textbf{Evaluation} & \textbf{Total} \\
\midrule
Temporal reasoning & 27 & 106 & 133 \\
Multi-session & 27 & 106 & 133 \\
Knowledge update & 15 & 63 & 78 \\
Single-session user & 14 & 56 & 70 \\
Single-session assistant & 11 & 45 & 56 \\
Single-session preference & 6 & 24 & 30 \\
\midrule
Total & 100 & 400 & 500 \\
\bottomrule
\end{tabularx}
\caption{LongMemEval-S question counts under the fixed stratified partition. Each question and its associated history are assigned together.}
\label{tab:longmemeval_split}
\end{table}

\subsection{Baselines and Implementations}
\label{app:baselines}
\paragraph{Shared evaluation interface.}
We reuse the baseline memory implementations through adapters that connect their construction and retrieval APIs to our standardized evaluation interface. The adapters map benchmark histories to each method's expected input format and return retrieved evidence to the shared QA model. Answer generation and judging follow the common benchmark protocol, with isolated memory state for each evaluation instance. Key configurations and integration adjustments are summarized below.

\paragraph{Full Context and RAG.}
Full Context supplies the conversation history within a 128,000-token budget, retaining the most recent messages in chronological order when truncation is needed. RAG retrieves the top-10 messages by embedding cosine similarity, preserving timestamps and speaker information, without query rewriting or reranking.

\paragraph{Memory system configurations.}
Mem0 uses batches of 20 messages for fact extraction and retrieves the top-10 facts with a threshold of 0.0. A-Mem retrieves the top-10 notes with linked-memory expansion. LightMem enables LLMlingua-2 compression and segmentation, uses \texttt{user\_only} messages and \texttt{flat} extraction, performs offline updates, and retrieves the top-20 memories. Its compression rate, extraction threshold, and update similarity threshold are 0.6, 0.1, and 0.9, respectively. SimpleMem uses 20-message windows with a two-message overlap, retrieval limits of 25, 5, and 5 for semantic, keyword, and structured search, and query planning with up to two reflection rounds. Zep uses Graphiti with retrieval limits of 20 for facts and entity summaries and \texttt{bge-reranker-v2-m3} reranking.

\paragraph{Integration adjustments.}
The integration layer handles model compatibility, usage accounting, and runtime error handling. Mem0 embedding inputs are truncated to 7,800 tokens when necessary; fact extraction still receives the full message batch. Mem0 and SimpleMem use minimal package initializers. Zep uses FalkorDB Lite instead of Neo4j, changing the storage and full-text search backend. These adjustments do not constitute a redesign of the baseline memory methods.

\subsection{Models and Evaluation Details}
\label{app:models_evaluation}

\paragraph{Serving models.}
We use GPT-4.1 mini and Qwen3-30B-A3B-Instruct-2507 as serving backbones for evaluation-set comparisons. Within each setting, the designated backbone handles both LLM-based memory operations and answer generation, and text-embedding-3-small provides embeddings. Qwen is served locally on eight NVIDIA H100 GPUs. Selected programs and baselines share the same designated models and benchmark-specific answering procedure. The QA prompt templates are provided in Appendix~\ref{app:qa_prompts}.

\paragraph{Accuracy.}
GPT-5 mini judges each predicted answer against the ground-truth answer using the benchmark-specific rubric. For reported benchmark results, we pool the resulting binary correctness judgments across all evaluated queries to obtain overall accuracy. The judge prompt templates are provided in Appendix~\ref{app:judge_prompts}. Reported accuracy gains and drops are absolute differences.

\paragraph{Generation settings.}
The shared QA wrapper sets a completion limit of 16,384 tokens and leaves temperature, top-p, and the generation seed unspecified, using API or serving defaults. Calls within memory programs can override wrapper defaults. The GPT-5 mini judge uses \texttt{low} reasoning effort and a 16,384-token completion limit. If a length-limit error occurs without visible output, the judge retries once with a 32,768-token limit.

\paragraph{Lifecycle serving cost.}
We record LLM input, LLM output, and embedding token usage across build, retrieve, compose, and answer, and apply the corresponding per-token prices of the designated models. Following \S\ref{ssec:cost}, we amortize the total cost over the answered queries for each history and take an unweighted mean across histories, reporting dollars per 1,000 queries. Judge and search-agent calls, general-purpose computation, and storage are excluded.

\paragraph{Token prices.}
We use fixed rates from our pricing configuration dated 21 September 2026, based on OpenAI\footnote{\url{https://openai.com/api/pricing/}} and Alibaba Cloud Model Studio\footnote{\url{https://www.alibabacloud.com/help/en/model-studio/model-pricing}} list prices. The rates below are in USD per million tokens.

\begin{center}
\small
\setlength{\tabcolsep}{3pt}
\begin{tabularx}{\columnwidth}{@{}Xrr@{}}
\toprule
\textbf{Model} & \textbf{Input} & \textbf{Output} \\
\midrule
GPT-4.1 mini & 0.40 & 1.60 \\
Qwen3-30B-A3B-Instruct-2507 & 0.20 & 0.80 \\
text-embedding-3-small & 0.02 & --- \\
\bottomrule
\end{tabularx}
\end{center}

For each call, we multiply input and output token counts by their respective rates and divide by one million. Embedding calls incur only the input charge. We apply no cached input discounts. Although Qwen runs locally, we value its token usage at the corresponding API rates. Reported costs therefore estimate serving cost at these fixed rates rather than actual infrastructure expenditure.

%% file: sections/appendix_supp.tex
\section{Supplementary Results and Analysis}
\label{app:supp}

\input{sections/appendix_transfer}

\input{sections/appendix_lineage}

\input{sections/appendix_retrieval}

\input{sections/appendix_evidence}

\input{sections/appendix_compose}

%% file: sections/appendix_transfer.tex
\subsection{Cross-Benchmark Transfer}
\label{app:transfer}
We evaluate the selected programs on the other benchmark without further evolution, using Qwen3-30B-A3B-Instruct-2507 as the serving backbone. Tables~\ref{tab:longmemeval_transfer} and~\ref{tab:locomo_transfer} report transfer from LoCoMo to LongMemEval and vice versa. Fixed input adapters reconcile the conversation formats before memory construction and retrieval, leaving the evolved code unchanged. QA, judging, embeddings, and cost accounting follow the target benchmark's evaluation protocol.

\paragraph{LoCoMo programs on LongMemEval.}
The adapter converts each dated user--assistant session into the two-speaker format expected by the LoCoMo programs. It preserves session and message order, reformats dates, and maps message roles to the speakers \texttt{User} and \texttt{Assistant}.

\input{sections/tab_longmemeval_transfer}

\paragraph{LongMemEval programs on LoCoMo.}
The reverse adapter converts each LoCoMo session into a dated message list, preserving order, speaker names, and image captions. For \sysname[-A\textsubscript{LME}], the first and second speakers map to \texttt{user} and \texttt{assistant}, respectively, to support its round grouping. \sysname[-P\textsubscript{LME}] does not use these roles and retains speaker names as role labels. No question date is supplied because LoCoMo does not provide one.

\input{sections/tab_locomo_transfer}

\paragraph{Serving-backbone transfer on LoCoMo.}
The temporal errors under backbone transfer help explain the ranking reversal reported in Section~\ref{ssec:overall_results}. \sysname[-A\textsubscript{LCM}] annotates relative time expressions with the conversation date, whereas \sysname[-P\textsubscript{LCM}] resolves them to absolute dates. Under Qwen, 134 of the former's 145 temporal errors return a conversation date rather than the event date. A concrete example is illustrated in Figure~\ref{fig:temporal_example}.

\input{sections/fig_temporal_example}

%% file: sections/tab_longmemeval_transfer.tex
\begin{table*}[!t]
\centering
\small
\setlength{\tabcolsep}{5pt}
\begin{tabularx}{\textwidth}{l >{\centering\arraybackslash}X >{\centering\arraybackslash}X >{\centering\arraybackslash}X >{\centering\arraybackslash}X >{\centering\arraybackslash}X >{\centering\arraybackslash}X >{\centering\arraybackslash}X c}
\toprule
\textbf{Method} & \textbf{Temp} & \textbf{Multi} & \textbf{Know} & \textbf{User} & \textbf{Asst} & \textbf{Pref} & \textbf{Avg} & \textbf{Cost} (\$/1k)$\downarrow$ \\
\midrule
\rowcolor{black!6}\multicolumn{9}{l}{\emph{Qwen3-30B-A3B-Instruct-2507}} \\
\midrule
Full Context & 43.40 & 41.51 & 74.60 & 91.07 & 93.33 & \underline{37.50} & 59.75 & 22.27 \\
RAG & 43.40 & \underline{53.77} & \textbf{85.71} & \underline{94.64} & \underline{97.78} & \textbf{41.67} & \textbf{66.00} & 2.62 \\
\midrule
Mem0 & 31.13 & 38.68 & 60.32 & 80.36 & 48.89 & \textbf{41.67} & 47.25 & 93.55 \\
A-Mem & 42.45 & 45.28 & 60.32 & 82.14 & \textbf{100.00} & \textbf{41.67} & 58.00 & 661.97 \\
LightMem & \underline{49.06} & \textbf{57.55} & 79.37 & \textbf{96.43} & 26.67 & \underline{37.50} & 59.50 & 26.59 \\
SimpleMem & 40.57 & 34.91 & 55.56 & 73.21 & 82.22 & 33.33 & 50.25 & 144.74 \\
Zep & 37.74 & 30.19 & 58.73 & 80.36 & 60.00 & 33.33 & 47.25 & 2367.28 \\
\midrule
\sysname[-A\textsubscript{LCM}] & \textbf{54.72} & 47.17 & \underline{84.13} & 92.86 & 95.56 & 25.00 & \underline{65.50} & 85.76 \\
\sysname[-P\textsubscript{LCM}] & 45.28 & 49.06 & \underline{84.13} & 91.07 & 95.56 & \textbf{41.67} & 64.25 & 69.17 \\
\bottomrule
\end{tabularx}
\caption{Cross-benchmark transfer from LoCoMo to LongMemEval without further evolution, using a fixed input adapter and the Qwen3-30B-A3B-Instruct-2507 serving backbone. Baseline results are reproduced from Table~\ref{tab:longmemeval}. Accuracy is reported in \%; Avg denotes overall accuracy across 400 questions, and cost is measured in dollars per 1,000 queries. Column abbreviations follow Table~\ref{tab:longmemeval}. The highest and second-highest accuracies in each column are \textbf{bold} and \underline{underlined}, respectively. Transfer details are provided in Appendix~\ref{app:transfer}.}
\label{tab:longmemeval_transfer}
\end{table*}

%% file: sections/tab_locomo_transfer.tex
\begin{table*}[!t]
\centering
\small
\setlength{\tabcolsep}{5pt}
\begin{tabularx}{\textwidth}{l >{\centering\arraybackslash}X >{\centering\arraybackslash}X >{\centering\arraybackslash}X >{\centering\arraybackslash}X >{\centering\arraybackslash}X >{\centering\arraybackslash}X}
\toprule
\textbf{Method} & \textbf{Multi} & \textbf{Open} & \textbf{Single} & \textbf{Temp} & \textbf{Avg} & \textbf{Cost} (\$/1k)$\downarrow$ \\
\midrule
\rowcolor{black!6}\multicolumn{7}{l}{\emph{Qwen3-30B-A3B-Instruct-2507}} \\
\midrule
Full Context & 58.16 & 53.01 & \textbf{89.41} & 31.78 & 70.01 & 7.07 \\
RAG & 27.62 & 38.55 & 58.05 & 16.67 & 43.08 & 0.14 \\
\midrule
Mem0 & 60.25 & 56.63 & 77.44 & 41.86 & 65.95 & 0.61 \\
A-Mem & 44.77 & 37.35 & 67.95 & 24.42 & 53.18 & 5.71 \\
LightMem & 54.81 & 45.78 & 76.20 & 28.68 & 60.98 & 0.48 \\
SimpleMem & \textbf{61.09} & 46.99 & \underline{86.52} & 35.66 & 69.32 & 3.24 \\
Zep & 50.21 & 44.58 & 57.22 & 30.62 & 49.89 & 6.31 \\
\midrule
\sysname[-A\textsubscript{LME}] & \underline{60.67} & \underline{62.65} & 86.24 & \textbf{65.12} & \textbf{75.90} & 1.94 \\
\sysname[-P\textsubscript{LME}] & 55.23 & \textbf{67.47} & 85.56 & \underline{64.34} & \underline{74.67} & 3.05 \\
\bottomrule
\end{tabularx}
\caption{Cross-benchmark transfer from LongMemEval to LoCoMo without further evolution, using a fixed input adapter and the Qwen3-30B-A3B-Instruct-2507 serving backbone. Baseline results are reproduced from Table~\ref{tab:locomo}. Accuracy is reported in \%; Avg denotes overall accuracy across 1,307 questions, and cost is measured in dollars per 1,000 queries. Column abbreviations follow Table~\ref{tab:locomo}. One context-overflow case for \sysname[-A\textsubscript{LME}] is counted as incorrect. The highest and second-highest accuracies in each column are \textbf{bold} and \underline{underlined}, respectively. Transfer details are provided in Appendix~\ref{app:transfer}.}
\label{tab:locomo_transfer}
\end{table*}

%% file: sections/fig_temporal_example.tex
\begin{figure*}[t]
\centering
\includegraphics[width=\textwidth]{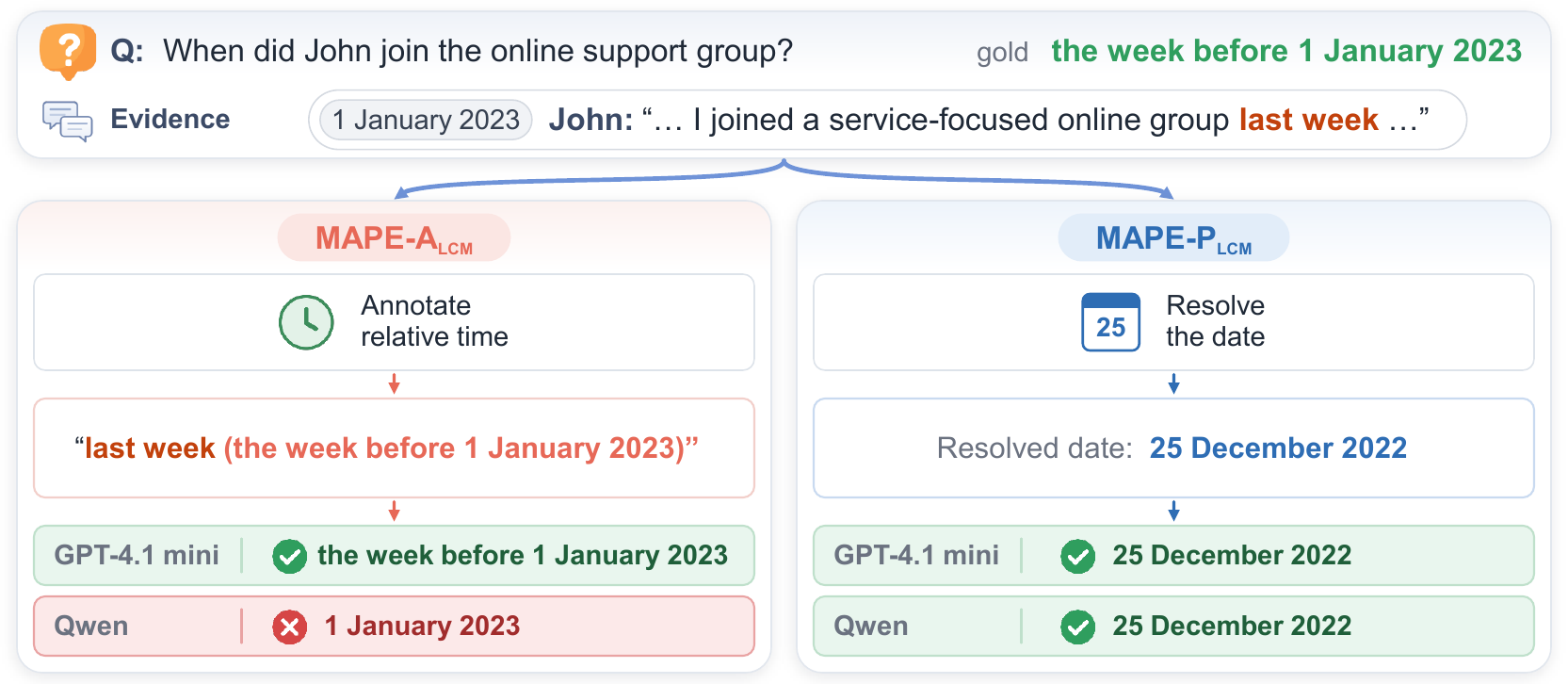}
\caption{A LoCoMo temporal example comparing \sysname[-A\textsubscript{LCM}] and \sysname[-P\textsubscript{LCM}] across serving backbones. The programs differ in the date representation and answer context passed to the QA model. Qwen answers are taken from held-out traces; GPT-4.1 mini answers are illustrative. Evidence and context excerpts are abbreviated.}
\label{fig:temporal_example}
\end{figure*}

%% file: sections/appendix_lineage.tex
\subsection{Search Trajectories and Lineage}
\label{app:lineage}
\input{sections/fig_lme_evolution}

Figure~\ref{fig:lme_evolution} shows how memory designs change during the two LongMemEval searches. Accuracy-only search introduces per-session note extraction at step 13 and retains it thereafter, increasing serving cost approximately sixfold. Accuracy--cost search switches to BM25-only retrieval from step 9 onward.

\begin{figure*}[t]
\centering
\includegraphics[width=\textwidth]{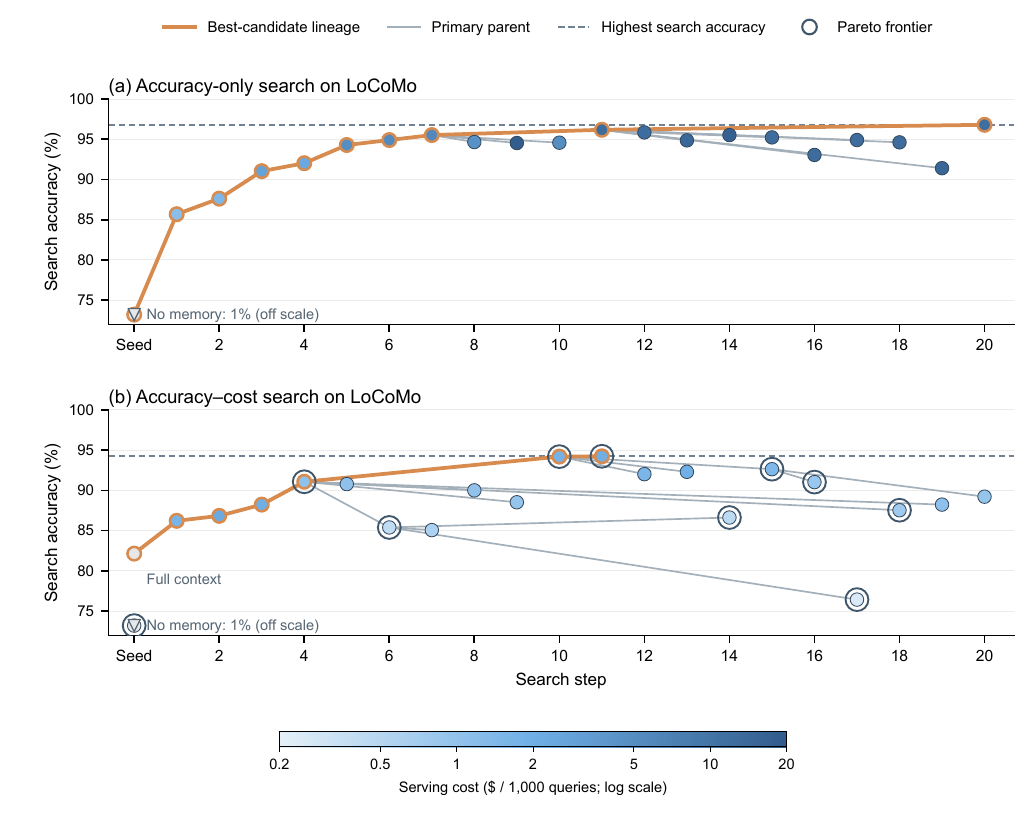}
\caption{LoCoMo search lineages using GPT-4.1 mini on the 233-question search set. Panels a and b show accuracy-only and accuracy--cost search, respectively. Nodes represent candidates; colour encodes serving cost on a logarithmic scale. Grey edges indicate primary parents, and orange paths trace the lineages of the highest-accuracy candidates. Dashed lines mark maximum search accuracy; dark rings mark the final Pareto frontier. The no-memory seed is displayed off-scale at the bottom.}
\label{fig:locomo_lineage}
\end{figure*}

Figure~\ref{fig:locomo_lineage} shows the candidate lineages on LoCoMo. Each edge connects a candidate to its first recorded parent. Additional parents consulted by the proposer are omitted. The orange paths trace the candidates with the highest search accuracy, whereas final programs are selected through the procedure described in Appendix~\ref{app:program_selection}. The selected \sysname[-A\textsubscript{LCM}] and \sysname[-P\textsubscript{LCM}] programs occur at steps 7 and 11, respectively.

%% file: sections/fig_lme_evolution.tex
\begin{figure*}[t]
\centering
\includegraphics[width=\textwidth]{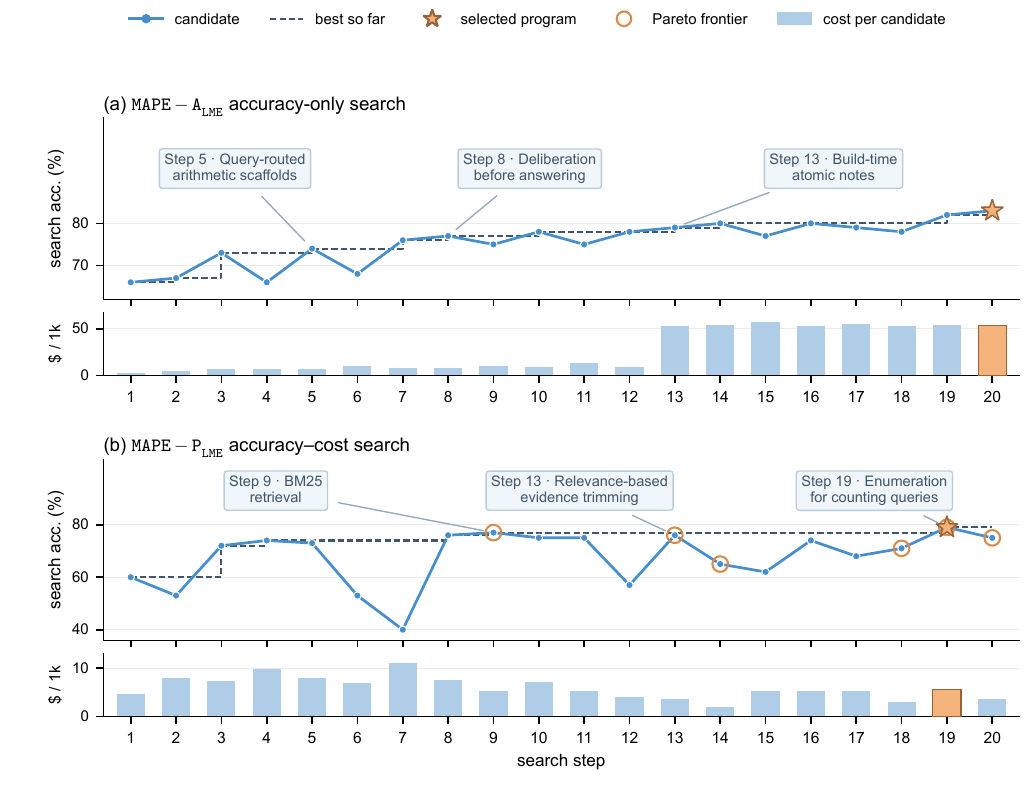}
\caption{LongMemEval search trajectories using Qwen3-30B-A3B-Instruct-2507 on the 100-question search set. Panels a and b show accuracy-only and accuracy--cost search, respectively. Blue lines show candidate accuracy, dashed lines show the best accuracy so far, and bars show lifecycle serving cost per 1,000 queries. Stars mark selected programs; orange circles mark the final Pareto frontier. Callouts identify representative design changes.}
\label{fig:lme_evolution}
\end{figure*}

%% file: sections/appendix_retrieval.tex
\subsection{Raw Conversation Rounds in \texorpdfstring{\sysname[-A\textsubscript{LME}]}{MAPE-A (LME)}}
\label{app:rawrounds}

\sysname[-A\textsubscript{LME}] returns extracted atomic notes alongside five retrieved raw conversation rounds, each consisting of a user message and the replies that follow it. To assess the contribution of raw evidence, we remove these rounds from the returned evidence while retaining them in the index and session scoring. All other components are kept unchanged. LoCoMo evaluation uses the fixed input adapter described in Appendix~\ref{app:transfer}.

\input{sections/tab_ablation_lme}
\input{sections/tab_ablation_locomo}

Removing raw rounds reduces overall accuracy by 6.50\% on LongMemEval and 12.17\% on LoCoMo, as shown in Tables~\ref{tab:ablation_lme} and~\ref{tab:ablation_locomo}. The largest decline on LongMemEval occurs for single-session-assistant questions. On LoCoMo, accuracy decreases across all categories, with the largest declines on temporal and single-hop questions. These results support providing raw conversational evidence alongside extracted notes.

%% file: sections/tab_ablation_lme.tex
\begin{table*}[!t]
\centering
\small
\setlength{\tabcolsep}{5pt}
\begin{tabularx}{\textwidth}{l >{\centering\arraybackslash}X >{\centering\arraybackslash}X >{\centering\arraybackslash}X >{\centering\arraybackslash}X >{\centering\arraybackslash}X >{\centering\arraybackslash}X >{\centering\arraybackslash}X}
\toprule
\textbf{Method} & \textbf{Temp} & \textbf{Multi} & \textbf{Know} & \textbf{User} & \textbf{Asst} & \textbf{Pref} & \textbf{Avg} \\
\midrule
{\fontfamily{lmtt}\bfseries MAPE-A\textsubscript{LME}} & \textbf{80.19} & \textbf{58.49} & \textbf{88.89} & \textbf{89.29} & \textbf{93.33} & \textbf{54.17} & \textbf{77.00} \\
\quad w/o raw rounds & 70.75 & 53.77 & 84.13 & 92.86 & 73.33 & 50.00 & 70.50 \\
\quad w/o RQS & 60.38 & 56.60 & 80.95 & 94.64 & 95.56 & 50.00 & 70.75 \\
\quad w/o both & 54.72 & 52.83 & 74.60 & 92.86 & 66.67 & 37.50 & 63.00 \\
\bottomrule
\end{tabularx}
\caption{Component ablation of \sysname[-A\textsubscript{LME}] on LongMemEval with the Qwen3-30B-A3B-Instruct-2507 serving backbone, providing the question-type breakdown of Table~\ref{tab:compose_ablation}. The bold row denotes the full system. \emph{W/o raw rounds} omits the five retrieved conversation rounds from the returned evidence while retaining them in the index; \emph{w/o RQS} bypasses composition and uses the default evidence rendering; \emph{w/o both} applies both ablations. Values are accuracies in \%, with Avg computed over all questions. Column abbreviations follow Table~\ref{tab:longmemeval}.}
\label{tab:ablation_lme}
\end{table*}

%% file: sections/tab_ablation_locomo.tex
\begin{table*}[!t]
\centering
\small
\setlength{\tabcolsep}{5pt}
\begin{tabularx}{\textwidth}{l >{\centering\arraybackslash}X >{\centering\arraybackslash}X >{\centering\arraybackslash}X >{\centering\arraybackslash}X >{\centering\arraybackslash}X}
\toprule
\textbf{Method} & \textbf{Multi} & \textbf{Open} & \textbf{Single} & \textbf{Temp} & \textbf{Avg} \\
\midrule
{\fontfamily{lmtt}\bfseries MAPE-A\textsubscript{LME}} & \textbf{60.67} & \textbf{62.65} & \textbf{86.24} & \textbf{65.12} & \textbf{75.90} \\
\quad w/o raw rounds & 49.37 & 56.63 & 74.00 & 50.39 & 63.73 \\
\quad w/o RQS & 58.16 & 50.60 & 90.51 & 36.82 & 71.46 \\
\quad w/o both & 52.30 & 49.40 & 77.03 & 38.37 & 63.12 \\
\bottomrule
\end{tabularx}
\caption{Component ablation of \sysname[-A\textsubscript{LME}] transferred to LoCoMo using the fixed input adapter described in Appendix~\ref{app:transfer} and the Qwen3-30B-A3B-Instruct-2507 serving backbone. Results provide the question-type breakdown of Table~\ref{tab:compose_ablation}. The bold row denotes the full system; ablation definitions follow Table~\ref{tab:ablation_lme}. Values are accuracies in \%, with Avg computed over all questions. Column abbreviations follow Table~\ref{tab:locomo}.}
\label{tab:ablation_locomo}
\end{table*}

%% file: sections/appendix_evidence.tex
\subsection{Retrieval Completeness and Error Attribution}
\label{app:evidence}

\input{sections/fig_evidence}

We use GPT-5 mini to assess evidence completeness in the analysed incorrect answers. The classifier receives the question, reference answer and evidence, model answer, and full retrieved context. It labels the retrieved evidence as complete, partial, or absent according to whether the context contains the information needed to derive the reference answer.

Figure~\ref{fig:evidence} combines partial and absent evidence into the incomplete category. This distinguishes errors accompanied by missing evidence from those that occur despite sufficient evidence. Separately, the classifier assesses whether further reasoning over the same context could plausibly recover the correct answer. The RQS transplantation described in Appendix~\ref{app:compose} corrects 42--60\% of errors judged recoverable, compared with 3--13\% of errors judged unrecoverable.

Figure~\ref{fig:evidence_acc} compares accuracy with estimated overall evidence completeness. Let $a$ denote answer accuracy and $r$ the proportion of incorrect answers with complete evidence. We estimate overall completeness as $\hat c = a + (1-a)r$, assuming complete evidence for every correct answer. This assumption can overestimate completeness. Because the estimate incorporates accuracy, we additionally examine the association between $r$ and accuracy across the six systems. Spearman's $\rho$ is 0.94 on LongMemEval and 0.77 on LoCoMo.

%% file: sections/fig_evidence.tex
\begin{figure*}[t]
\centering
\includegraphics[width=\textwidth]{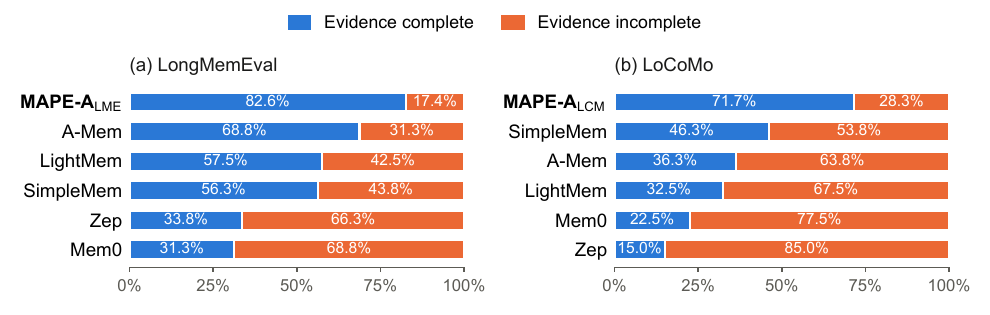}
\caption{Evidence completeness among incorrect answers with the Qwen3-30B-A3B-Instruct-2507 serving backbone. Panels a and b show LongMemEval and LoCoMo, respectively. Each bar sums to 100\% of the analysed errors for one system. Blue denotes complete retrieved evidence despite an incorrect answer; orange denotes incomplete evidence. Classification details are provided in Appendix~\ref{app:evidence}.}
\label{fig:evidence}
\end{figure*}

\begin{figure*}[t]
\centering
\includegraphics[width=\textwidth]{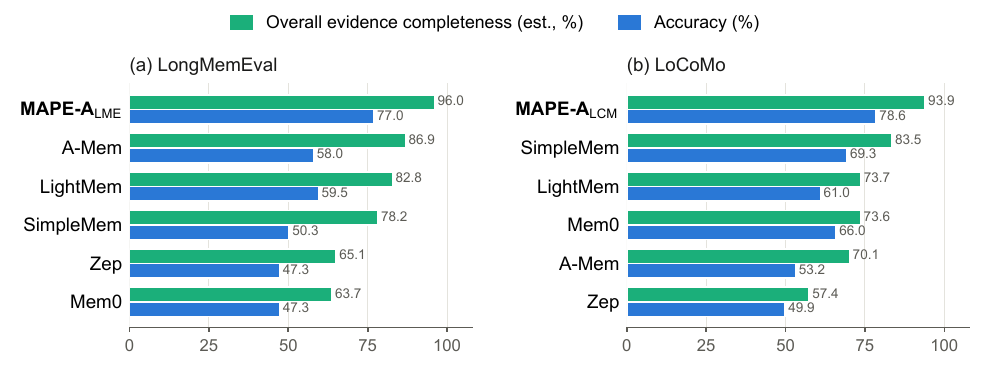}
\caption{Estimated overall evidence completeness and answer accuracy on LongMemEval in panel a and LoCoMo in panel b. Systems are ordered by estimated completeness. The estimate combines accuracy with the error labels shown in Figure~\ref{fig:evidence}, assuming complete evidence for every correct answer. The calculation and correlation analysis are detailed in Appendix~\ref{app:evidence}.}
\label{fig:evidence_acc}
\end{figure*}

%% file: sections/appendix_compose.tex
\subsection{Query-Routed Scaffolding: Design, Ablation, and Transplant}
\label{app:compose}

\paragraph{Design.}
Query-Routed Scaffolding (RQS) implements the compose stage of \sysname[-A\textsubscript{LME}], as shown in Figure~\ref{fig:memorph_a}. It processes retrieved evidence through three steps: routing, scaffolding, and deliberation.

\textit{Route.} Rules based on the query wording assign each question to one of six intents: temporal, aggregation, difference, current-value, preference, or direct lookup. The assigned intent determines the subsequent processing strategy.

\textit{Scaffold.} For temporal questions, the scaffold extracts event dates and explicitly stated durations. It anchors relative expressions such as ``last week'' to the relevant session date and resolves them through date arithmetic. For aggregation questions, it extracts matching items from each session, removes duplicates, and counts or sums the items in code. Two independent extraction passes are compared. If they disagree, both item lists are retained for deliberation. Difference questions use extracted values for programmatic subtraction, while current-value questions select the most recently dated value. Preference questions use relevant user details to guide personalised responses. Direct lookups proceed to deliberation without a specialised computational scaffold.

\textit{Deliberate.} A single LLM call applies intent-specific instructions to the scaffold outputs and retrieved evidence, with session notes ordered chronologically. The resulting reasoning and supporting quote are combined with the evidence in the composed context and passed to the shared QA model for final answer generation. Deliberation can revise scaffold outputs against the evidence. All LLM calls within RQS use the designated serving backbone, which also performs QA. The QA model and answering prompt are held fixed across compared systems.

\paragraph{Evaluation protocol.}
We isolate the contribution of RQS by replaying the same recorded retrieval for each paired comparison. Memory construction, retrieved evidence, and QA settings are held fixed. The ablation compares the full system with a variant that replaces RQS with the benchmark's default evidence rendering. We evaluate \sysname[-A\textsubscript{LME}] on LongMemEval and on LoCoMo through the fixed input adapter described in Appendix~\ref{app:transfer}.

\paragraph{Transplantation.}
We apply the same RQS module to SimpleMem, LightMem, A-Mem, Zep, and Mem0 while retaining each system's memory construction and retrieval. A fixed adapter maps retrieved memory units to notes, groups them by available dates, and removes exact duplicates within each group. Method-specific fields are normalized to this common representation; units without temporal metadata are placed in an undated session. Each system is compared with default evidence rendering under the same QA configuration.

\input{sections/tab_compose_lme}

\paragraph{Results.}
Removing RQS reduces overall accuracy by 6.25\% on LongMemEval and 4.44\% on LoCoMo, with the largest declines on temporal questions, as shown in Tables~\ref{tab:ablation_lme} and~\ref{tab:ablation_locomo}. Transplantation improves all five memory systems by 2.75--9.25\% on LongMemEval, as shown in Figure~\ref{fig:compose_transfer}. Table~\ref{tab:compose_lme} provides the question-type breakdown. Gains vary across categories; on LoCoMo, removing RQS improves single-hop accuracy despite reducing overall performance.

%% file: sections/tab_compose_lme.tex
\begin{table*}[!t]
\centering
\small
\setlength{\tabcolsep}{5pt}
\begin{tabularx}{\textwidth}{ll >{\centering\arraybackslash}X >{\centering\arraybackslash}X >{\centering\arraybackslash}X >{\centering\arraybackslash}X >{\centering\arraybackslash}X >{\centering\arraybackslash}X >{\centering\arraybackslash}X}
\toprule
\textbf{Method} & \textbf{RQS} & \textbf{Temp} & \textbf{Multi} & \textbf{Know} & \textbf{User} & \textbf{Asst} & \textbf{Pref} & \textbf{Avg} \\
\midrule
SimpleMem & without & 39.62 & 34.91 & 52.38 & 67.86 & 82.22 & 41.67 & 49.25 \\
 & with & 56.60 & 38.68 & 60.32 & 76.79 & 84.44 & 58.33 & 58.50 \\
 & $\Delta$ & $+16.98$ & $+3.77$ & $+7.94$ & $+8.93$ & $+2.22$ & $+16.67$ & $+9.25$ \\
\midrule
LightMem & without & 50.00 & 58.49 & 79.37 & 94.64 & 26.67 & 29.17 & 59.25 \\
 & with & 73.58 & 57.55 & 90.48 & 91.07 & 26.67 & 58.33 & 68.25 \\
 & $\Delta$ & $+23.58$ & $-0.94$ & $+11.11$ & $-3.57$ & $0.00$ & $+29.17$ & $+9.00$ \\
\midrule
A-Mem & without & 36.79 & 44.34 & 61.90 & 80.36 & 97.78 & 37.50 & 55.75 \\
 & with & 54.72 & 50.00 & 66.67 & 82.14 & 95.56 & 41.67 & 63.00 \\
 & $\Delta$ & $+17.92$ & $+5.66$ & $+4.76$ & $+1.79$ & $-2.22$ & $+4.17$ & $+7.25$ \\
\midrule
Zep & without & 35.85 & 31.13 & 58.73 & 76.79 & 62.22 & 33.33 & 46.75 \\
 & with & 52.83 & 37.74 & 52.38 & 80.36 & 60.00 & 33.33 & 52.25 \\
 & $\Delta$ & $+16.98$ & $+6.60$ & $-6.35$ & $+3.57$ & $-2.22$ & $0.00$ & $+5.50$ \\
\midrule
Mem0 & without & 28.30 & 41.51 & 60.32 & 80.36 & 48.89 & 45.83 & 47.50 \\
 & with & 33.96 & 49.06 & 57.14 & 78.57 & 46.67 & 50.00 & 50.25 \\
 & $\Delta$ & $+5.66$ & $+7.55$ & $-3.17$ & $-1.79$ & $-2.22$ & $+4.17$ & $+2.75$ \\
\bottomrule
\end{tabularx}
\caption{RQS transplantation into five human-designed memory systems on the LongMemEval evaluation set with the Qwen3-30B-A3B-Instruct-2507 serving backbone. Results provide the question-type breakdown of the overall gains shown in Figure~\ref{fig:compose_transfer}. For each system, \emph{without} uses the default evidence rendering, while \emph{with} applies RQS to the same retrieved evidence. Values are accuracies in \%; $\Delta$ denotes the absolute accuracy change from adding RQS in \%. Deltas are computed before rounding. Avg is computed over all questions, and column abbreviations follow Table~\ref{tab:longmemeval}.}
\label{tab:compose_lme}
\end{table*}

%% file: sections/appendix_pending.tex
\section{Prompt Templates}
\label{app:prompts}

We present the QA and LLM-as-a-judge prompts used to generate and score answers on LoCoMo and LongMemEval. Placeholders are filled with the corresponding context, question, and evaluation fields at runtime.

\subsection{QA Prompt Templates}
\label{app:qa_prompts}

We use consistent QA prompt templates across all evaluated memory systems. For LoCoMo, Prompt~\ref{prompt:qa_locomo} is used for multi-hop, open-domain, and single-hop questions, while Prompt~\ref{lst:qa_locomo_temporal} is used for temporal questions.

For LongMemEval, we use Prompt~\ref{lst:qa_longmemeval} for all questions, supplying the question date alongside the context. Across systems, the context is produced by the memory pipeline, while the answering instructions remain fixed within each benchmark and question type.

\begin{figure*}
\begin{promptbox}[label={prompt:qa_locomo}]{LoCoMo QA}
Based on the context: {context}, write an answer in the form of a short phrase for the following question. Answer with exact words from the context whenever possible.

Question: {query}
Short answer:
\end{promptbox}
\end{figure*}

\begin{figure*}    
\begin{promptbox}[label={lst:qa_locomo_temporal}]{LoCoMo QA: Temporal Questions}
Based on the context: {context}, answer the following question. Use DATE of CONVERSATION to answer with an approximate date. Please generate the shortest possible answer, using words from the conversation where possible, and avoid using any subjects.

Question: {query}
Short answer:
\end{promptbox}
\end{figure*}

\begin{figure*}[t]
\begin{promptbox}[label={lst:qa_longmemeval}]{LongMemEval QA}
I will give you several history chats between you and a user. Please answer the question based on the relevant chat history. Answer concisely (a short phrase or 1 sentence) using exact wording from the chats when possible.

History Chats:
{history}

Current Date: {question_date}
Question: {query}
Answer:
\end{promptbox}
\end{figure*}

\subsection{LLM-as-a-Judge Prompt Templates}
\label{app:judge_prompts}

The judge receives the question, reference answer, and model prediction, and returns a binary correctness score with a brief explanation. Scores are aggregated into accuracy using the evaluation protocol described in Appendix~\ref{app:models_evaluation}.

Prompt~\ref{lst:judge_locomo} evaluates all LoCoMo questions and accepts equivalent temporal expressions. LongMemEval uses separate rubrics for temporal reasoning, knowledge updates, and user preferences, as shown in Prompts~\ref{lst:judge_longmemeval_temporal}, \ref{lst:judge_longmemeval_knowledge}, and~\ref{lst:judge_longmemeval_preference}. Prompt~\ref{lst:judge_longmemeval_other} covers the remaining question types. These rubrics account for differences in valid answers, including updated information and personalised responses.

\begin{figure*}[t]
\begin{promptbox}[label={lst:judge_locomo}]{LoCoMo Judge}
Your task is to label an answer to a question as CORRECT (1) or WRONG (0). You will be given the following data:
    (1) a question (posed by one user to another user),
    (2) a 'gold' (ground truth) answer,
    (3) a generated answer
which you will score as CORRECT (1) or WRONG (0).

The point of the question is to ask about something one user should know about the other user based on their prior conversations.
The gold answer will usually be a concise and short answer that includes the referenced topic, for example:
Question: Do you remember what I got the last time I went to Hawaii?
Gold answer: A shell necklace
The generated answer might be much longer, but you should be generous with your grading - as long as it touches on the same topic as the gold answer, it should be counted as CORRECT (1).

For time related questions, the gold answer will be a specific date, month, year, etc. The generated answer might be much longer or use relative time references (like "last Tuesday" or "next month"), but you should be generous with your grading - as long as it refers to the same date or time period as the gold answer, it should be counted as CORRECT (1). Even if the format differs (e.g., "May 7th" vs "7 May"), consider it CORRECT if it's the same date.

Now it's time for the real question:
Question: {query}
Gold answer: {reference}
Generated answer: {prediction}

First decide CORRECT or WRONG. Output ONLY a JSON object:
{"reason": "<one sentence explanation>", "score": <1 if CORRECT else 0>}
\end{promptbox}
\end{figure*}

\begin{figure*}[t]
\begin{promptbox}[label={lst:judge_longmemeval_temporal}]{LongMemEval Judge: Temporal Reasoning}
I will give you a question, a correct answer, and a response from a model. Please answer yes if the response contains the correct answer. Otherwise, answer no. If the response is equivalent to the correct answer or contains all the intermediate steps to get the correct answer, you should also answer yes. If the response only contains a subset of the information required by the answer, answer no. In addition, do not penalize off-by-one errors for the number of days. If the question asks for the number of days/weeks/months, etc., and the model makes off-by-one errors (e.g., predicting 19 days when the answer is 18), the model's response is still correct.

Question: {query}
Correct answer: {reference}
Model response: {prediction}

Decide yes or no. Output ONLY a JSON object:
{"reason": "<one sentence explanation>", "score": <1 if yes else 0>}
\end{promptbox}
\end{figure*}

\begin{figure*}[t]
\begin{promptbox}[label={lst:judge_longmemeval_knowledge}]{LongMemEval Judge: Knowledge Update}
I will give you a question, a correct answer, and a response from a model. Please answer yes if the response contains the correct answer. Otherwise, answer no. If the response contains some previous information along with an updated answer, the response should be considered as correct as long as the updated answer is the required answer.

Question: {query}
Correct answer: {reference}
Model response: {prediction}

Decide yes or no. Output ONLY a JSON object:
{"reason": "<one sentence explanation>", "score": <1 if yes else 0>}
\end{promptbox}
\end{figure*}

\begin{figure*}[t]
\begin{promptbox}[label={lst:judge_longmemeval_preference}]{LongMemEval Judge: Single-Session Preference}
I will give you a question, a rubric for desired personalized response, and a response from a model. Please answer yes if the response satisfies the desired response. Otherwise, answer no. The model does not need to reflect all the points in the rubric. The response is correct as long as it recalls and utilizes the user's personal information correctly.

Question: {query}
Correct answer: {reference}
Model response: {prediction}

Decide yes or no. Output ONLY a JSON object:
{"reason": "<one sentence explanation>", "score": <1 if yes else 0>}
\end{promptbox}
\end{figure*}

\begin{figure*}[t]
\begin{promptbox}[label={lst:judge_longmemeval_other}]{LongMemEval Judge: Other Questions}
I will give you a question, a correct answer, and a response from a model. Please answer yes if the response contains the correct answer. Otherwise, answer no. If the response is equivalent to the correct answer or contains all the intermediate steps to get the correct answer, you should also answer yes. If the response only contains a subset of the information required by the answer, answer no.

Question: {query}
Correct answer: {reference}
Model response: {prediction}

Decide yes or no. Output ONLY a JSON object:
{"reason": "<one sentence explanation>", "score": <1 if yes else 0>}
\end{promptbox}
\end{figure*}

\clearpage
\onecolumn
\subsection{Evolutionary Search Prompts}
\label{app:evolution_prompts}
We present selected excerpts from the proposer and diagnosis prompt templates used in the evolutionary search. Wording is preserved, with line wrapping adjusted and omissions marked by \texttt{[...]}. The proposer excerpts are grouped by role and objective, interface and constraints, proposal strategy, and output requirements. Group headings are editorial. The two objective blocks are alternatives selected by the search setting. Template placeholders and escaped braces are retained. The diagnosis output instructions are placed last for readability.

\begin{promptbox}[label={prompt:proposer_excerpts},listing options={basicstyle=\ttfamily\footnotesize,columns=fullflexible,keepspaces=true,breaklines=true,breakindent=0pt,showstringspaces=false,literate={—}{{\textemdash}}1 {•}{{\textbullet}}1}]{Proposer: Selected Excerpts}
ROLE AND SEARCH OBJECTIVE
You are a researcher designing memory systems for AI agents.
[...]
How it does that — what it stores, in what structure, and how it finds things again — is yours to decide.

Accuracy-only:
WHAT COUNTS AS A BETTER DESIGN, on this run: a higher `accuracy_<dataset>` than the best harness already has, compared at the same `stage_<dataset>`. That is the entire criterion.

Accuracy--cost:
WHAT COUNTS AS A BETTER DESIGN, on this run: a candidate that NO existing harness beats on BOTH axes at once — one that would join the Pareto frontier. Pushing EITHER edge outward counts, and they count equally: more accurate at the same cost, the same accuracy for fewer tokens, or a trade-off nobody occupies yet.
[...]
INTERFACE AND CONSTRAINTS
Class inheriting from `forge.memo_class.MemoClass`, implementing the two required hooks, and optionally the third:
  async def build_memory_from_data(self, recorder) -> None
  async def retrieve_memory_for_query(self, recorder) -> Dict
  async def compose_memory_context(self, recorder, query, retrieved, qa_metadata)  # optional
[...]
Every LLM or embedding call inside your harness goes through `common.llm.Agent` / `common.llm.Embedding` with `model` LEFT UNSET, so it uses the run's unified models.
[...]
DO NOT hardcode shortcuts that game search-set scores.
[...]
PRIOR EXPERIENCE AND PROPOSAL STRATEGY
You decide which priors to study and how deeply.
[...]
• INCREMENTAL — find specific weakness in prior candidates (failure patterns in traces, weak QA category, brittle retrieval, missing dispatch, etc.) and target it with a focused change.
• EXPLORATORY — try a fundamentally different design philosophy, possibly one inspired by a research line you investigated via WebSearch/WebFetch. The structure can share little with prior best candidates.
[...]
Substantive over cosmetic. Whichever stance you take, the change should be real — not a prompt rewording, constant tweak, or rename.
[...]
REQUIRED OUTPUT
Write to `{new_dir_rel}/`:
  1. memo.py               new MemoClass subclass (+ src/ if it needs one).
  2. meta.json             {{"parent_ids":     [...],
                             "description":    "...",
                             "design_choices": [{{"choice": "...",
                                                  "how":    "..."}}, ...]}}
[...]
Each entry's `how` covers BOTH mapping (code mechanism) AND rationale (why this choice over alternatives + what failure mode it targets).
\end{promptbox}

\clearpage
\noindent The diagnosis agent attributes findings to build, retrieve, compose, answer, judge, or data, and uses unattributed when the evidence is insufficient. Its core analysis and reporting instructions are shown below.

\begin{promptbox}[label={prompt:diagnosis_excerpts},listing options={basicstyle=\ttfamily\small,columns=fullflexible,keepspaces=true,breaklines=true,breakindent=0pt,showstringspaces=false,literate={—}{{\textemdash}}1}]{Diagnosis: Selected Excerpts}
You are handed ONE harness that has finished evaluation. Say what happened and why, from the evidence, and leave the next proposer something it could not have worked out on its own.

You do not propose a design. You do not write harness code.
[...]
## Method

1. **Read the hypothesis, then the diff.** What did this candidate change relative to its parent, mechanically — not what its docstring claims.
2. **Read the numbers on every axis this run records**, against the parent. `metrics.json` splits the dollars by phase, so a cost change can be placed in build, retrieve or answer.
3. **Attribute the failures.** Go through the wrong answers. Group them, then for each group follow the evidence to the earliest operation that made the outcome inevitable. Cite what you read: a `qa` record, a `call` record, a line of the harness's code.
4. **Write the leads.** What the evidence says is worth trying next, each one attached to the finding that motivates it.
[...]
If the evidence does not settle it, use **unattributed** and say what would settle it (a level the run did not record, a call that was truncated).
[...]
Exactly two files, both under `diagnostics/`, and nothing else anywhere:

    diagnostics/<<ANALYST_ID>>.md      prose, for the next proposer to read
    diagnostics/<<ANALYST_ID>>.json    the same findings, structured
\end{promptbox}